\documentclass[11pt]{article}

\usepackage[final]{acl}

\usepackage{times}
\usepackage{latexsym}

\usepackage[T1]{fontenc}

\usepackage[utf8]{inputenc}

\usepackage{microtype}

\usepackage{inconsolata}

\usepackage{graphicx}

\usepackage{hyperref}
\usepackage{url}

\usepackage{booktabs}
\usepackage{adjustbox}
\usepackage{amsmath}
\usepackage{amssymb}
\usepackage{algpseudocode}
\usepackage{algorithm}
\usepackage{newfloat}
\usepackage{listings}
\usepackage{graphicx}
\usepackage{multirow}

\usepackage[many]{tcolorbox}
\tcbuselibrary{listings, breakable, skins}

\newtcblisting{promptbox}{
    listing only,
    listing options={
        basicstyle=\ttfamily\footnotesize,
        breaklines=true,
        columns=fullflexible
    },
    colback=gray!5,
    colframe=gray!70,
    boxrule=0.5pt,
    arc=2pt,
    left=4pt,
    right=4pt,
    top=2pt,
    bottom=2pt,
    grow to left by=10pt,
    breakable
}
\usepackage{subcaption}
\usepackage{xcolor}

\title{HiCoLoRA: Addressing Context-Prompt Misalignment via Hierarchical Collaborative LoRA for Zero-Shot DST}

\author{%
    Shuyu Zhang\textsuperscript{\rm 1}\thanks{These authors contributed equally to this work.},\;
    Yifan Wei\textsuperscript{\rm 2 *},\;
    Xinru Wang\textsuperscript{\rm 3},\;
    Yanmin Zhu\textsuperscript{\rm 1}\thanks{Corresponding Author.}, \; \\
    \textbf{Yangfan He}\textsuperscript{\rm 4},\;
    \textbf{Yixuan Weng}\textsuperscript{\rm 5},\; 
    \textbf{Bin Li}\textsuperscript{\rm 6 \dag},\; 
    \textbf{Yujie Liu}\textsuperscript{\rm 7}\; 
    \\
    \textsuperscript{\rm 1}Shanghai Jiao Tong University, 
    \textsuperscript{\rm 2}Beihang University, 
    \textsuperscript{\rm 3}University of Sydney, \\
    \textsuperscript{\rm 4}University of Minnesota Twin Cities,  
    \textsuperscript{\rm 5}Westlake University,\\
    \textsuperscript{\rm 6}SIAT, CAS, 
    \textsuperscript{\rm 7}Beijing Institute of Graphic Communication\\ 
    \small{
   \href{mailto:carsonz@sjtu.edu.cn}{carsonz@sjtu.edu.cn},
   \href{mailto:yzhu@cs.sjtu.edu.cn}{yzhu@cs.sjtu.edu.cn},
   \href{mailto:b.li2@siat.ac.cn}{b.li2@siat.ac.cn},
 }
}

\begin{document}
\maketitle
\begin{abstract}

Zero-shot Dialog State Tracking (zs-DST) is essential for enabling Task-Oriented Dialog Systems (TODs) to generalize to new domains without costly data annotation. A central challenge lies in the semantic misalignment between dynamic dialog contexts and static prompts, leading to inflexible cross-layer coordination, domain interference, and catastrophic forgetting. To tackle this, we propose Hierarchical Collaborative Low-Rank Adaptation (HiCoLoRA\footnote{https://github.com/carsonz/HiCoLoRA}), a framework that enhances zero-shot slot inference through robust prompt alignment. It features a hierarchical LoRA architecture for dynamic layer-specific processing (combining lower-layer heuristic grouping and higher-layer full interaction), integrates Spectral Joint Domain-Slot Clustering to identify transferable associations (feeding an Adaptive Linear Fusion Mechanism), and employs Semantic-Enhanced SVD Initialization (SemSVD-Init) to preserve pre-trained knowledge. Experiments on multi-domain datasets MultiWOZ and SGD show that HiCoLoRA outperforms baselines, achieving SOTA in zs-DST.

\end{abstract}

\section{Introduction}

\begin{figure}[t]
\begin{center}
    \includegraphics[width=0.95\linewidth, keepaspectratio]{"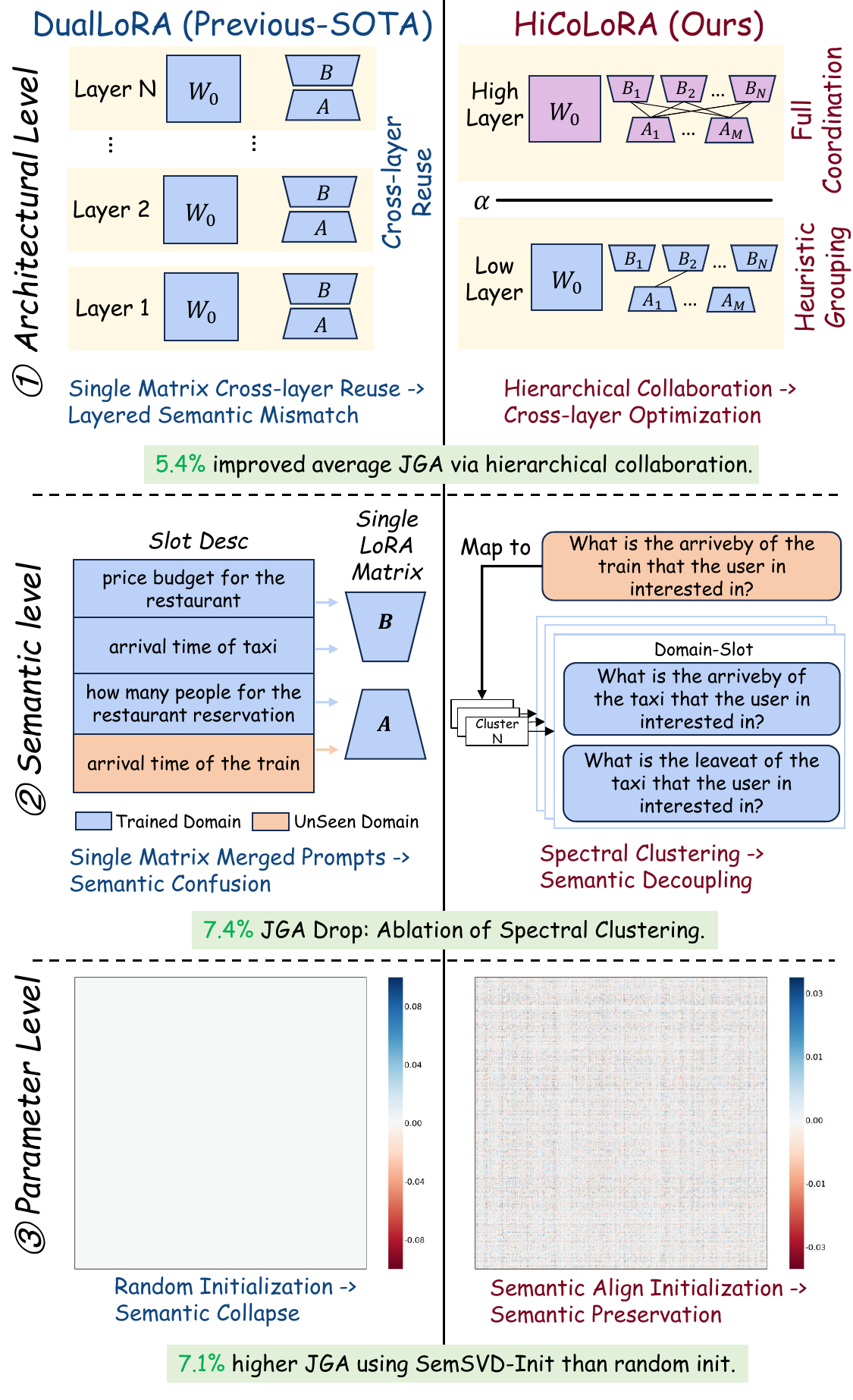"}
\end{center}
\caption{Three critical challenges motivating our work: (1) Architectural rigidity hinders cross-layer coordination in Transformers, limiting fine-grained semantic alignment; (2) Coupling of domain-shared and domain-specific semantics causes cross-domain confusion; (3) Random parameter initialization distorts pre-trained knowledge, exacerbating catastrophic forgetting.}
\label{fig:motivation}
\end{figure}

Task-Oriented Dialog Systems (TODs) assist users to complete specific tasks, such as hotel or flight bookings \cite{luo2024zero, niu2024enhancing}. To generalize to new domains without extensive training, TODs rely on zero-shot Dialog State Tracking (zs-DST), yet face challenges of data scarcity and domain adaptation. A central obstacle is the semantic misalignment between dynamic dialog contexts and static prompts. This misalignment limits effective cross domain transfer.

To address data scarcity and enable generalization, existing approaches for zs-DST include data augmentation \cite{he2025simulation}, prompt engineering \cite{liu2025towards, wang2024can, aksu2023prompter}, and parameter-efficient fine-tuning (PEFT). Among PEFT methods, Low-Rank Adaptation (LoRA) \cite{wang2024instruct, occhipinti2024fine} has gained prominence for zs-DST \cite{yi2025intent, aksu2023prompter} due to its ability to adapt large models with minimal parameters. However, zs-DST faces three intrinsic challenges: \textbf{(1)} zero-shot semantic misalignment, where dynamic dialog contexts must align with static prompts under domain shift; \textbf{(2)} domain interference, where shared and specific semantics become entangled; and \textbf{(3)} irreversible knowledge forgetting, where scarce target data cannot recover distorted pre-trained knowledge. While recent multi-LoRA variants such as DualLoRA \cite{luo2024zero}, CoLA \cite{zhou2025cola}, HydraLoRA \cite{tian2024hydralora}, and MTL-LoRA \cite{yang2025mtl} enhance adaptability, they treat all transformer layers uniformly, failing to leverage the inherent hierarchy where lower layers capture local semantic atoms and higher layers integrate global intents \cite{liu-etal-2024-fantastic}. This structural limitation exacerbates the above challenges, motivating our proposed hierarchical collaborative design.

These limitations stem from a structural mismatch between dynamic dialog contexts and static prompts as illustrated in Fig.~\ref{fig:motivation}, which manifests in three critical research questions:
\textbf{(RQ1)} Rigid hierarchical designs hinder effective cross-layer weight sharing, limiting fine-grained semantic alignment in deeper layers.
\textbf{(RQ2)} A single adaptation matrix conflates domain-agnostic and domain specific signals, causing semantic confusion between domains.
\textbf{(RQ3)} The use of random initialization for LoRA parameters can distort pre-trained knowledge and exacerbate catastrophic forgetting.

To address the three limitations, we propose Hierarchical Collaborative Low-Rank Adaptation (HiCoLoRA), a novel framework inspired by DualLoRA's prompt augmentation \cite{luo2024zero} and CoLA's multi-LoRA grouping \cite{zhou2025cola}. Departing from ``uniform layer processing'', our contributions are:
\textbf{(1)} A Hierarchical Collaborative Architecture with lower-layer heuristic grouping and higher-layer full interaction, resolving RQ1 via dynamic cross-layer coordination;
\textbf{(2)} Spectral Joint Clustering and Adaptive Fusion disentangling domain-shared and specific semantics addressing RQ2;
\textbf{(3)} Semantic-Enhanced SVD Initialization preserving pre-trained knowledge against RQ3.

Extensive experiments on MultiWOZ and SGD multi-domain datasets demonstrate that HiCoLoRA significantly outperforms previous SOTA methods, validating the effectiveness of our proposed framework in addressing the core challenges of zs-DST.

\section{Related Work}

\noindent \textbf{Layer-Specific Algorithms in Transformers.}
The hierarchical processing in Transformers is well established: lower layers capture local semantic atoms, while higher layers integrate them into global intent representations~\cite{liu-etal-2024-fantastic,wang2025hierarchicalreasoningmodel}. To exploit this asymmetry, various layer specific algorithms have been proposed, including hierarchical LoRA~\cite{xiao2024hivg,guo2024fedhlt}, dynamic layer replacement~\cite{xiong2024efficient}, attention head pruning~\cite{he2025olica,zayed2024fairness}, and split attention mechanisms~\cite{lin2025hsplitlora}. While these methods improve efficiency by leveraging unequal layer contributions, they typically treat layers uniformly, lacking mechanisms to coordinate their distinct semantic roles, particularly crucial for dynamic dialog contexts where fine grained alignment between evolving utterances and static prompts is required. HiCoLoRA addresses this gap through explicit hierarchical collaboration, enabling dynamic cross-layer coordination.

\noindent \textbf{PEFT with LoRA.}
LoRA and its variants have become prominent for zs-DST due to their parameter efficiency~\cite{zhang2025parameter,liu2025adaptive,jabbarvaziri2025parameter}. DualLoRA~\cite{luo2024zero} uses dual adapters to align contexts and prompts, while multi-adapter approaches like HydraLoRA~\cite{tian2024hydralora}, CoLA~\cite{zhou2025cola}, and MTL-LoRA~\cite{yang2025mtl} enhance cross-task collaboration. Methods like RoSA~\cite{nikdan2024rosa} combine low-rank and sparse adaptations for efficiency, while initialization strategies like PiSSA~\cite{meng2024pissa} and MiLoRA~\cite{zhang2024milora} aim to better preserve pre-trained knowledge. Spectral based adaptations have also been explored~\cite{zhang2024spectral,wang2024lora}. Yet, these approaches still struggle with the semantic misalignment between dynamic dialog contexts and static slot prompts, often conflating domain agnostic and domain specific signals, which hinders zero-shot generalization. HiCoLoRA directly tackles this via spectral joint clustering and adaptive fusion, enabling disentangled and aligned representations.

\noindent \textbf{zs-DST and Goal Accuracy.}
zs-DST aims to generalize to unseen domains without annotated data. Early methods like TRADE~\cite{wu2019transferable} and SUMBT~\cite{lee2019sumbt} relied on task-specific architectures. With the advent of PLMs, generation based approaches like SimpleTOD~\cite{hosseini2020simple} and intent enhanced methods~\cite{yi2025intent} have advanced the field. Prompt based methods like Prompter~\cite{aksu2023prompter} and dual adapter approaches like DualLoRA~\cite{luo2024zero} further improve cross-domain transfer, while synthetic data methods like LUAS~\cite{niu2024enhancing} address data scarcity. However, these approaches remain limited by rigid layer processing, insufficient semantic disentanglement, and knowledge distortion during adaptation. HiCoLoRA fundamentally optimizes these challenges through its hierarchical cross-layer coordination and spectral domain-slot disentanglement.

\section{Method}
\label{sec:method}

\begin{figure*}[!ht]
\begin{center}
    \includegraphics[width=0.9\textwidth, keepaspectratio]{"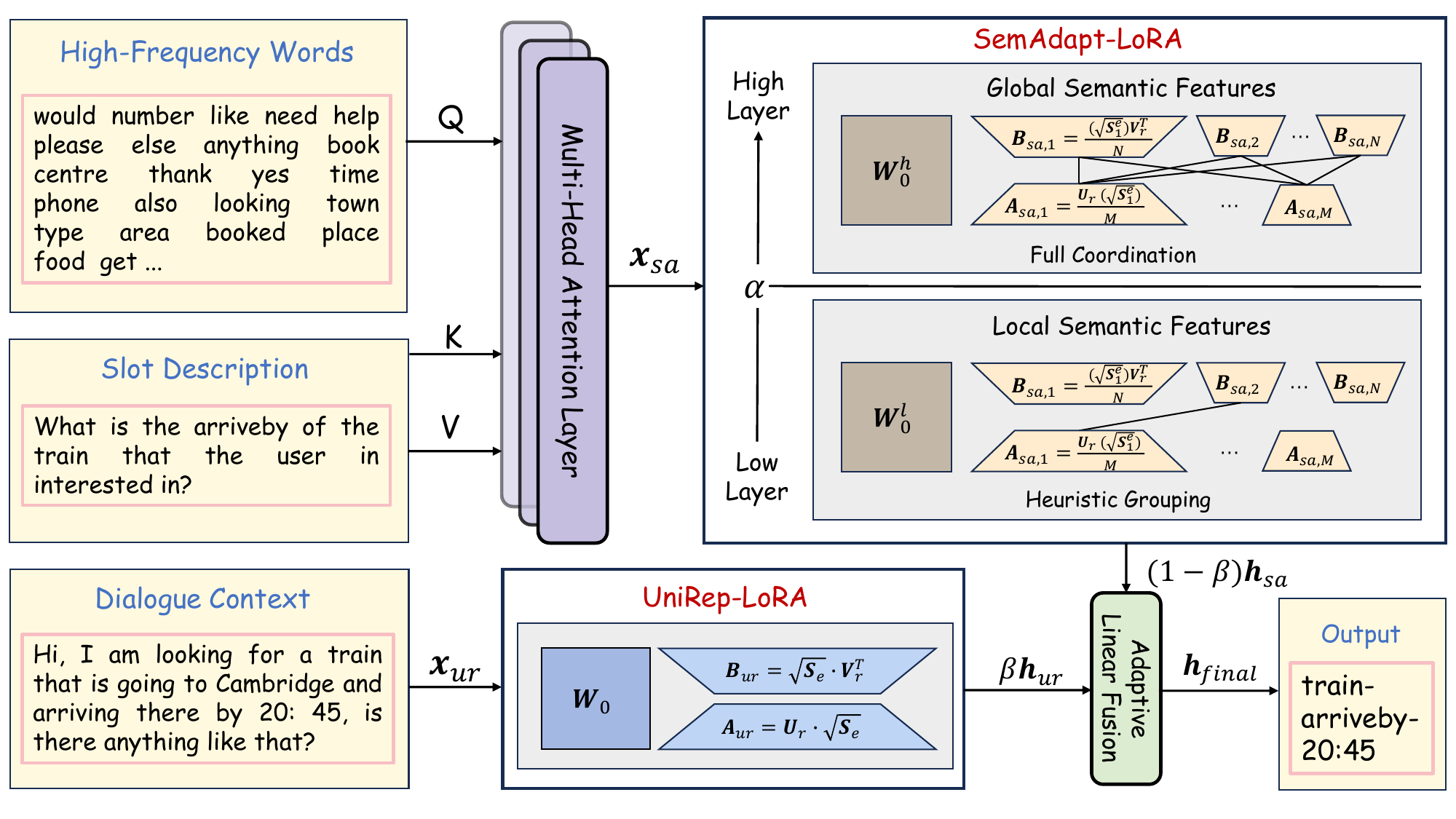"}
\end{center}
\caption{The HiCoLoRA framework combines: (1) UniRep-LoRA and SemAdapt-LoRA with Adaptive Linear Fusion balancing domain-agnostic and domain-specific features; (2) Spectral Joint Domain-Slot Clustering disentangling domain semantics to guide fusion; (3) SemSVD-Init preserving pre-trained knowledge via singular value modulation. These synergistically address context-prompt misalignment, enhancing zero-shot slot inference.}
\label{fig:flowchart}
\end{figure*}

We propose Hierarchical Collaborative Low-Rank Adaptation (HiCoLoRA, Fig.~\ref{fig:flowchart}) to address the research questions: A hierarchical collaborative architecture resolves cross-layer rigidity (RQ1) by enabling dynamic layer specific coordination; Spectral joint domain-slot clustering that disentangles domain shared and domain specific semantics (RQ2), guiding adaptive fusion between domain-agnostic (UniRep-LoRA) and domain specific (SemAdapt-LoRA) representations; and Semantic Enhanced SVD Initialization (SemSVD-Init) that preserves pre-trained knowledge against catastrophic forgetting (RQ3). These components systematically mitigate context prompt misalignment and strengthen zero-shot generalization.

\subsection{Design Rationale Overview}
\label{sec:design_rationale}

HiCoLoRA decomposes the context-prompt misalignment in zs-DST into three semantic flow challenges, each addressed by a dedicated component:

\textbf{(1) Hierarchical Collaboration (Sec.~\ref{sec:UniRep-LoRA} \& \ref{sec:SemAdapt-LoRA})} targets \textit{cross-layer semantic flow}. Lower layers preserve local semantic atoms (e.g., entity mentions) via heuristic grouping; higher layers integrate them into global intent chains via full collaboration. This mirrors the natural Transformer hierarchy and resolves rigid layer uniformity (RQ1).

\textbf{(2) Spectral Joint Clustering + Adaptive Fusion (Sec.~\ref{sec:fusion_mechanism} \& \ref{sec:clustering})} disentangles \textit{domain-shared vs. domain-specific semantics}. Clustering identifies transferable slot associations; adaptive gating dynamically balances universal and specialized representations, preventing semantic conflation (RQ2).

\textbf{(3) SemSVD-Init (Sec.~\ref{sec:SemSVD-Init})} preserves \textit{pre-trained semantic structure}. By modulating singular values according to domain-slot relevance, it initializes adaptation from principal semantic directions rather than noise, avoiding catastrophic forgetting (RQ3).

These components collectively convert the static prompt-dynamic context gap into a structured, layer-aware, and semantically disentangled learning problem.

\subsection{Problem Formulation}
\label{sec:problem_formulation}

\textbf{Dialogue State Tracking.} Given a dialogue history $C_t = \{u_1, u_2, \dots, u_t\}$ at turn $t$ and a schema prompt $\mathcal{P}$ that describes target slots and their candidate values, the goal of Dialogue State Tracking is to predict the current dialogue state $S_t$, a set of slot-value pairs that captures the user's goals. Formally, we learn a parameterized function $f_\theta$ such that:
\begin{equation}
S_t = \arg\max_{S} P(S \mid C_t, \mathcal{P}; \theta),
\end{equation}
where $\theta$ denotes the trainable parameters of the model. In our framework, the input is processed through two streams: a universal context representation $\boldsymbol{x}_{ur}$ derived from $C_t$, and a semantic-aligned representation $\boldsymbol{x}_{sa}$ obtained via attention interaction between $C_t$ and $\mathcal{P}$. The final prediction is generated from the fused hidden state $\boldsymbol{h}_{final}$ over the frozen pre-trained weights $\boldsymbol{W}_0$.

\textbf{Zero-Shot Setting.} Let $\mathcal{D}_{\text{all}}$ denote the set of all available domains. We partition $\mathcal{D}_{\text{all}}$ into two disjoint sets: the \textit{seen} domain set $\mathcal{D}_{\text{seen}}$ for training and the \textit{unseen} domain set $\mathcal{D}_{\text{unseen}}$ for evaluation, satisfying $\mathcal{D}_{\text{seen}} \cup \mathcal{D}_{\text{unseen}} = \mathcal{D}_{\text{all}}$ and $\mathcal{D}_{\text{seen}} \cap \mathcal{D}_{\text{unseen}} = \emptyset$. The model is trained exclusively on samples from $\mathcal{D}_{\text{seen}}$ and evaluated on $\mathcal{D}_{\text{unseen}}$ without any fine-tuning on target domain data. This strict separation aligns with our experimental protocol in Section 4.1 (e.g., excluding Police and Hospital domains in MultiWOZ 2.1).

\textbf{Semantic Clustering.} To facilitate knowledge transfer across disjoint domains, we employ Spectral Joint Clustering (Section 3.4) to identify transferable semantic associations. This yields domain clusters $\mathcal{D}^M$ and slot prompt clusters $\mathcal{X}^N$, which guide the adaptive fusion between domain-agnostic (UniRep-LoRA) and domain-specific (SemAdapt-LoRA) representations, enabling robust generalization under the zero-shot constraint.

\subsection{Universal Representation LoRA (UniRep-LoRA)}
\label{sec:UniRep-LoRA}

UniRep-LoRA is designed to efficiently capture domain-agnostic semantic information from the dialog context $\boldsymbol{x}_{ur}$, such as universal slots for time and location. By freezing the parameters of the pre-trained model $\boldsymbol{W}_0$ and updating only the low-rank matrices $\boldsymbol{B}_{ur}$ and $\boldsymbol{A}_{ur}$:
\begin{equation}
\boldsymbol{h}_{ur} = \boldsymbol{W}_0 x_{ur} + \boldsymbol{B}_{ur} \boldsymbol{A}_{ur} \boldsymbol{x}_{ur} \label{eq:unirep_lora}.
\end{equation}

UniRep-LoRA and SemAdapt-LoRA are combined via adaptive linear fusion, balancing general and domain-specific representations to mitigate context-prompt misalignment for zero-shot scenarios.

\subsection{Semantic Adaptation LoRA (SemAdapt-LoRA)}
\label{sec:SemAdapt-LoRA}
Unlike UniRep-LoRA which captures universal features, SemAdapt-LoRA specializes in domain specific prompt optimization, dynamically adjusting their influence across domains (addressing RQ2). 
We first employ a Multi-Head Attention module to enhance semantic alignment between dialog contexts and slot descriptions: high frequency dialog words from the training set serve as $\boldsymbol{Q}$, while slot descriptions act as $\boldsymbol{K}$ and $\boldsymbol{V}$, yielding the aligned representation $\boldsymbol{x}_{sa}$. 
This allows different heads to capture diverse semantic correlations, providing richer local semantic inputs for subsequent hierarchical processing.

To enable fine grained adaptation, we introduce two sets of trainable low-rank matrices: 
$\boldsymbol{A}_{sa}^{m}|_{m=1}^M$ for domain common prompt encoding, and $\boldsymbol{B}_{sa}^{n}|_{n=1}^N$ for cluster specific domain-slot reconstruction. 
Here, $M$ denotes the number of identified domain clusters, and $N$ the number of semantic clusters for slot prompts. 
$\boldsymbol{A}_{sa}^{m}$ compresses high dimensional prompt semantics into a low-rank space, while $\boldsymbol{B}_{sa}^{n}$ reconstructs domain specific representations from these compressed features.

To ensure effective collaboration between $\boldsymbol{A}_{sa}^{m}$ and $\boldsymbol{B}_{sa}^{n}$ across layers (addressing RQ1), we design a cross-layer collaborative module that respects the distinct semantic roles of Transformer layers: 
lower layers capture \textit{local semantic features} serving as semantic atoms, while higher layers model \textit{global semantic features} and guide lower-layer feature extraction via attention based suppression of irrelevant associations. 
This hierarchical approach moves beyond uniform layer processing, forming a semantic chain from local cues to global intent.

\noindent\textbf{Heuristic Grouping.}
For lower layers that encode local semantic atoms, we adopt heuristic grouping to aggregate semantically similar parameters.
The selection of domain-specific and slot-specific matrices is performed via a differentiable weighted aggregation:
\begin{multline}
\boldsymbol{h}_{sa} = \boldsymbol{W}_0^l \boldsymbol{x}_{sa} + N \cdot \left( \sum_{n=1}^{N} \beta_n \boldsymbol{B}_{sa}^n \right) \\
\cdot M \cdot \left( \sum_{m=1}^{M} \alpha_m \boldsymbol{A}_{sa}^m \right) \boldsymbol{x}_{sa},
\label{eq:heuristic_grouping}
\end{multline}
where \(\alpha_m\) and \(\beta_n\) are attention weights computed as the cosine similarity between the input \(\boldsymbol{x}_{sa}\) and the cluster centroids of domains \(\mathcal{D}^M\) and slot prompts \(\mathcal{X}^N\), respectively.
During training, the weights are obtained via Gumbel-Softmax for differentiable selection; at inference, softmax is used for efficiency.
The scalars \(N\) and \(M\) normalize the expected update magnitude, ensuring that the gradient scale in sparse heuristic grouping remains comparable to that in the dense full collaboration of higher layers (Eq.~\ref{eq:full_collaboration}).

\noindent\textbf{Full Collaboration.} 
Higher layers integrate all local semantic atoms through full collaboration, capturing implicit cross slot associations, such as \textit{train-arriveby} and \textit{destination}. This process can suppress irrelevant signals via attention guidance from lower layers:
\begin{equation}
\boldsymbol{h}_{sa} = \boldsymbol{W}_0^h \boldsymbol{x}_{sa} + \sum_{n=1}^N \boldsymbol{B}_{sa}^n \sum_{m=1}^M \boldsymbol{A}_{sa}^m \boldsymbol{x}_{sa},
\label{eq:full_collaboration}
\end{equation}
where $\boldsymbol{W}_0^h$ denotes higher-layer weights.

\subsection{Adaptive Linear Fusion Mechanism}
\label{sec:fusion_mechanism}
To dynamically integratethe two LoRA modules, we introduce a gated fusion mechanism with a learnable coefficient $\beta$ (trained end-to-end):
\begin{equation}
\boldsymbol{h}_{final} = \beta \boldsymbol{h}_{ur}+ (1-\beta) \boldsymbol{h}_{sa}, \quad \beta \in (0,1).
\end{equation}
Unlike static weighting, this adaptive gating allows the model to adjust the contribution of each module per turn based on the dialog context and slot descriptions, thereby mitigating the dynamic static prompt misalignment inherent in zs-DST.

\subsection{Spectral Clustering of Domains and Slot Prompts}
\label{sec:clustering}

We propose a spectral joint clustering mechanism to capture semantic relatedness across domains and slot prompts, addressing the transferable associations needed for zero-shot adaptation. Domains often share abstract categories, such as \textit{train} and \textit{taxi} as transportation, while slot prompts are formatted as structured pairs \textit{\{domain-slot: question\}}, for instance \textit{\{train-arriveby: what is the arrival time?\}}. This reveals cross-domain semantic commonalities, such as \textit{train-arriveby} and \textit{taxi-arriveby} both expressing temporal attributes.

To cluster these semantically, we encode domain names and slot prompts using a T5 encoder into dense vector representations. Spectral clustering is then applied via Laplacian matrix eigendecomposition. The optimal cluster numbers ($M$ for domains and $N$ for slot prompts) are determined by maximizing the silhouette coefficient, resulting in clusters $\mathcal{D}^M$ and $\mathcal{X}^N$.

To ensure strict zero-shot compliance, all clustering centroids and hyperparameters ($M$, $N$) are determined exclusively on training domains schemas. Target domain schemas are never used in clustering; they are projected onto the fixed source centroids via cosine similarity at inference.

\subsection{Semantic Enhanced SVD Initialization (SemSVD-Init)}
\label{sec:SemSVD-Init}

While standard LoRA employs a statistically controlled initialization ($\boldsymbol{A} \sim \mathcal{N}, \boldsymbol{B} = \boldsymbol{0}$), PiSSA~\citep{meng2024pissa} indicates this yields semantically unaligned gradient directions ($\nabla \boldsymbol{A} \to \boldsymbol{0}, \nabla \boldsymbol{B} \to \text{Random}$), which is suboptimal for zero-shot scenarios with scarce data. Our SemSVD-Init addresses this by modulating singular values according to domain-slot semantics. Notably, SemSVD-Init preserves the pre-trained model's initial output via residual adjustment ($\boldsymbol{W}_{res} = \boldsymbol{W}_0 - \boldsymbol{B}\boldsymbol{A}$), ensuring no knowledge distortion at inference start, while enabling adaptation from principal semantic directions rather than noise.

Formally, we first perform SVD on the pre-trained weight $\boldsymbol{W}_0$:
\begin{equation}
\boldsymbol{W}_0 = \boldsymbol{U}_r \boldsymbol{\Sigma}_r \boldsymbol{V}_r^T.
\end{equation}
A correlation matrix $\boldsymbol{R}$ is computed by cosine similarity between the right singular vectors $\boldsymbol{V}_r$ and the cluster embeddings $\text{T5}_{\text{en}}(\mathcal{X}^N)$:
\begin{equation}
\boldsymbol{R} = \cos(\boldsymbol{V}_r, \mathrm{T5}_{en}(\mathcal{X}^N)).
\end{equation}
Using these correlations, the singular values are enhanced based on maximum category relevance for each vector:
\begin{multline}
\boldsymbol{S}_e = \mathrm{diag}\left( \sigma_1 \cdot \mathrm{ReLU}(1+\lambda \boldsymbol{R}_1), \ldots, \right. \\
\left. \sigma_r \cdot \mathrm{ReLU}(1+\lambda \boldsymbol{R}_r) \right),
\end{multline}
where $\boldsymbol{R}_k$ is the relevance score for the $k$-th singular vector, $\mathrm{ReLU}(x)=\max(0,x)$, and $\lambda$ is a hyperparameter. The LoRA matrices are initialized as:
\begin{equation}
\begin{split}
\boldsymbol{A}_{ur} &= \sqrt{\boldsymbol{S}_e} \boldsymbol{V}_r^T, \\
\boldsymbol{B}_{ur} &= \boldsymbol{U}_r \sqrt{\boldsymbol{S}_e}.
\end{split}
\end{equation}
Finally, the residual weight matrix $\boldsymbol{W}_{res}$ is adjusted to preserve key knowledge of the pre-trained model:
\begin{equation}
\boldsymbol{W}_{res} = \boldsymbol{W}_{0} -  \boldsymbol{B}_{ur}\boldsymbol{A}_{ur}.
\end{equation}

\section{Experiments}
\subsection{Experimental Setup}
\noindent \textbf{Dataset.}
We evaluate on two standard multi-domain TOD benchmarks: MultiWOZ 2.1 and Schema-Guided Dialog (SGD). Both are split with strict domain separation for zero-shot evaluation (details in Appendix~\ref{sec:dataset_stats}).

\noindent \textbf{Base Model.}
Following previous SOTA PEFT baselines (e.g., DualLoRA \cite{luo2024zero}, Prompter \cite{aksu2023prompter}), we adopt \textbf{PPTOD-small} \cite{su_multi-task_2022} as our primary backbone to ensure fair comparison. Unlike vanilla T5, PPTOD is a unified plug-and-play model specifically pre-trained for TOD. Its dialogue multi-task pre-training strategy allows the model to learn primary TOD task completion skills from heterogeneous dialog corpora. This specialized pre-training equips PPTOD with stronger prior knowledge for TOD sub-tasks compared to general-purpose PLMs, making it particularly suitable for zero-shot DST where domain-specific cues are scarce.

\noindent \textbf{Baseline.}
To evaluate the generalizability of the proposed HiCoLoRA method, we conduct a comparison against representative baselines and SOTA approaches with details in Appendix~\ref{sec: baseline}. Additionally, comparisons with recent advanced LoRA variants and larger scale LLMs are included to thoroughly assess scalability and generalization.

\noindent \textbf{Metrics.}
We evaluate all models using Joint Goal Accuracy (JGA) and Average Goal Accuracy (AGA). JGA measures the rate of turns with all slots exactly matched, indicating system-level reliability. AGA calculates the ratio of correctly predicted to total slots, accounting for missed true slots and errors, reflecting fine-grained slot recall and local semantic alignment.
The metrics' formulas and additional experimental details are provided in Appendices~\ref{sec:formula} and~\ref{sec:exp_impl_details}.

\subsection{Main Results and Analysis}

We summarize the main findings from Table \ref{table: MultiWOZ_modified} (MultiWOZ) and Appendix Table \ref{table: SGD_modified} (SGD) below:

\begin{table*}[t]
    \centering
    {\fontsize{9}{11}\selectfont 
    \resizebox{0.9\textwidth}{!}{
        \begin{tabular}{@{}ccccccccc@{}}
            \toprule
            \textbf{Method} & \textbf{Year} & \textbf{Base Model} & \textbf{Attraction} & \textbf{Hotel} & \textbf{Restaurant} & \textbf{Train} & \textbf{Taxi} & \textbf{Average} \\ \midrule
            TRADE & 2019 & customized seq2seq & 20.1 & 14.2 & 12.6 & 22.4 & 59.2 & 25.7 \\
            MA-DST & 2020 & TRADE & 22.5 & 16.3 & 13.6 & 22.8 & 59.3 & 26.9 \\
            SUMBT & 2019 & BERT-base & 22.6 & 19.1 & 16.5 & 22.5 & 59.5 & 28.0 \\
            GPT2-DST & 2021 & GPT2-base & 23.7 & 18.5 & 21.1 & 24.3 & 59.1 & 29.3 \\
            T5DST & 2021 & T5-small & 31.9 & 20.7 & 20.1 & 28.8 & 64.1 & 33.1 \\
            SlotDM-DST & 2022 & T5-small & 33.9 & 18.9 & 20.8 & 37.0 & 66.3 & 35.4 \\
            T5DST* & 2021 & PPTOD-small & 35.5 & 20.0 & 25.3 & 35.3 & 65.6 & 36.4 \\
            Prompter & 2023 & PPTOD-small & 35.8 & 19.2 & 26.0 & 39.0 & 66.3 & 37.2 \\
            DCC & 2023 & T5-small & 35.8 & \textbf{24.8} & 22.9 & 40.2 & 65.9 & 37.9 \\
            DualLoRA (Prev. SOTA)  & 2024 & PPTOD-small & 37.1 & 18.9 & 27.9 & 42.4 & 67.2 & 38.7 \\
            \textbf{HiCoLoRA (Ours)} & 2025 & PPTOD-small & $\mathbf{38.86\pm0.19}$ & $20.41\pm0.11$ & $30.96\pm0.21$ & $\mathbf{44.49\pm0.31}$ & $\mathbf{68.91\pm0.34}$ & $\mathbf{40.8\pm0.18}$ \\
            \cmidrule(lr){1-9}
            \textbf{\% Gain vs DualLoRA} &  &  & \textbf{+4.9} & \textbf{+7.9} & \textbf{+11.1} & \textbf{+5.9} & \textbf{+2.1} & \textbf{+5.4} \\
            \bottomrule
        \end{tabular}
    }
    }
    \caption{Zero-shot JGA (\%) on the MultiWOZ dataset with relative improvement over previous SOTA. All results of baselines were reported from original papers. T5DST* was excerpted from Prompter \cite{aksu2023prompter}.}
    \label{table: MultiWOZ_modified}
\end{table*}

\noindent \textbf{Overall Performance Superiority.}
HiCoLoRA achieves new SOTA results on both MultiWOZ and SGD, with an average JGA of 40.8 on MultiWOZ and consistent gains across all SGD domains. This improvement stems from our hierarchical adaptation design, which overcomes three key limitations of prior work: rigid feature engineering in traditional methods, catastrophic forgetting in full fine-tuning, and shallow or uniform adaptation in recent SOTA models. Notably, HiCoLoRA attains an AGA of 93.8\% on SGD Trains, demonstrating its ability to preserve rare slot semantics through SemSVD-Init and maintain layer wise specificity.

\noindent \textbf{Component-Wise Efficacy Validation.}
HiCoLoRA performs robustly across diverse domain types, attributable to its tailored architectural components. In \textbf{transfer-rich domains} such as \textit{Media}, it achieves 75.9\% JGA, a 9.4\% improvement over DualLoRA, owing to spectral clustering that identifies cross domain commonalities and disentangles domain shared semantics. In \textbf{domain-specific regimes} such as \textit{Hotel}, the model attains 20.4\% JGA (+7.9\%), where SemSVD-Init preserves sparse slot semantics otherwise distorted by random initialization. For \textbf{context-sensitive domains} like \textit{Messaging}, adaptive fusion dynamically balances static prompts against volatile dialog contexts, yielding a 4.0\% gain over DualLoRA's static weighting.

\noindent \textbf{Architectural Validation Against Prev. SOTA.}
The hierarchical design of HiCoLoRA directly addresses core limitations of DualLoRA. \textbf{Cross-Layer Rigidity (RQ1)}: DualLoRA's uniform processing hinders fine-grained alignment. HiCoLoRA's heuristic grouping (lower layers) and full collaboration (higher layers) enable dynamic coordination, boosting \textit{Restaurant} JGA to +11.1\%. \textbf{Semantic Conflation (RQ2)}: Where DualLoRA's single adaptation matrix confuses domain signals, spectral joint clustering separates transport domain semantics (\textit{Taxi}: 44.9 JGA, +2.1\% error reduction). \textbf{Knowledge Distortion (RQ3)}: DualLoRA's random initialization loses rare slot knowledge. SemSVD-Init preserves pre-trained semantics, critical for \textit{Flights}' technical terms JGA +8.1\%.

\noindent \textbf{Discussion.}
HiCoLoRA fundamentally resolves context prompt misalignment via hierarchical adaptation, spectral semantic disentanglement, and knowledge preserving initialization. By overcoming DualLoRA's structural limitations, our method establishes a new paradigm for zs-DST. Future work will address extreme sparse slots through domain aware initialization refinements.

\subsection{Ablation Study}

We conduct an ablation study (Table \ref{table: ablation} in Appendex~\ref{sec: ablation_table}) to assess the contribution of each key component of HiCoLoRA.

\noindent \textbf{w/o Swap Hierarchical Strategies} swapping layer-wise strategies, using heuristic grouping in high layers and full collaboration in low layers. This variant sees an 8.3\% drop in the average JGA. The decline arises because it disrupts synergy: lower layers are designed to capture local semantic atoms, while higher layers model global intents. Swapping strategies break this division, validating the assumption that layer-specific roles are critical for performance.

\noindent \textbf{w/o Adaptive Linear Fusion} replacing adaptive gating with DualLoRA's static $\beta=0.5$, causing a 12.0\% JGA drop, notably in Attraction and Train domains. This exacerbates that static weighting cannot dynamically balance UniRep-LoRA (domain-agnostic) and SemAdapt-LoRA (domain-specific) features across layers. Unlike the adaptive mechanism that mitigates cross-layer semantic mismatches, static $\beta$ locks in misalignment, leading to performance drops.

\noindent \textbf{w/o Spectral Joint Cluster} discarding spectral clustering, retaining the same number of $M$ and $N$ but without identifying transferable domain-slot associations. Its average JGA drops 7.4\%, notably in Train and Taxi domains. The decline occurs because spectral clustering captures cross-domain semantic commonalities, such as ``arriveby'' in trains and taxis sharing temporal attributes, to guide effective feature fusion. Without it, the model fails to leverage transferable associations, weakening the alignment between domain-slot prompts and dynamic contexts, thus hindering zero-shot generalization.

\noindent \textbf{w/ Kaiming Init} using Kaiming initialization for matrix $\boldsymbol{A}$ and zero initialization for matrix $\boldsymbol{B}$ results in a 6.6\% decline in the average JGA. SemSVD-Init preserves pre-trained semantics by modulating singular values, thereby suppressing catastrophic forgetting. Without this mechanism, random initialization induces knowledge distortion and forgetting, preventing the model from retaining critical semantics and impairing its zero-shot transfer capability.

\noindent \textbf{w/ PiSSA Init} using PiSSA initialization, trailing HiCoLoRA by 4.7\% but outperforming random init. PiSSA partially addresses RQ3 but not as effectively: it retains pre-trained knowledge but lacks alignment of singular values to domain-slot semantics, limiting performance.

\noindent \textbf{w/ MiLoRA Init} using MiLoRA initialization, resulting in a significant performance drop. This degradation occurs because the MiLoRA strategy, which is designed to update minor singular components, is misaligned with the limited parameter capacity and the flat singular value spectrum of the T5-small model. Consequently, it fails to preserve crucial pre-trained semantics and severely impairs the model's zero-shot transfer capability.

Ablation studies demonstrate that the hierarchical collaborative architecture, adaptive fusion, spectral clustering, and SemSVD-Init components of HiCoLoRA are all indispensable. These components synergistically address the three core research questions, outperform baselines in zs-DST, and thus validate the efficacy of the proposed design.

\section{Analysis}

This section evaluates HiCoLoRA design choices to validate its mechanisms, including rank sensitivity, high layer ratio, and attention alignment (Figs. \ref{fig: rank}--\ref{fig: attn}), examining expressiveness balance, semantic flow optimization, and sustained attention for zero-shot performance.


\subsection{Model Mechanism Analysis}

\begin{figure}[t]
    \centering
    \begin{minipage}[t]{0.48\textwidth}
        \centering
        \includegraphics[width=\linewidth, keepaspectratio]{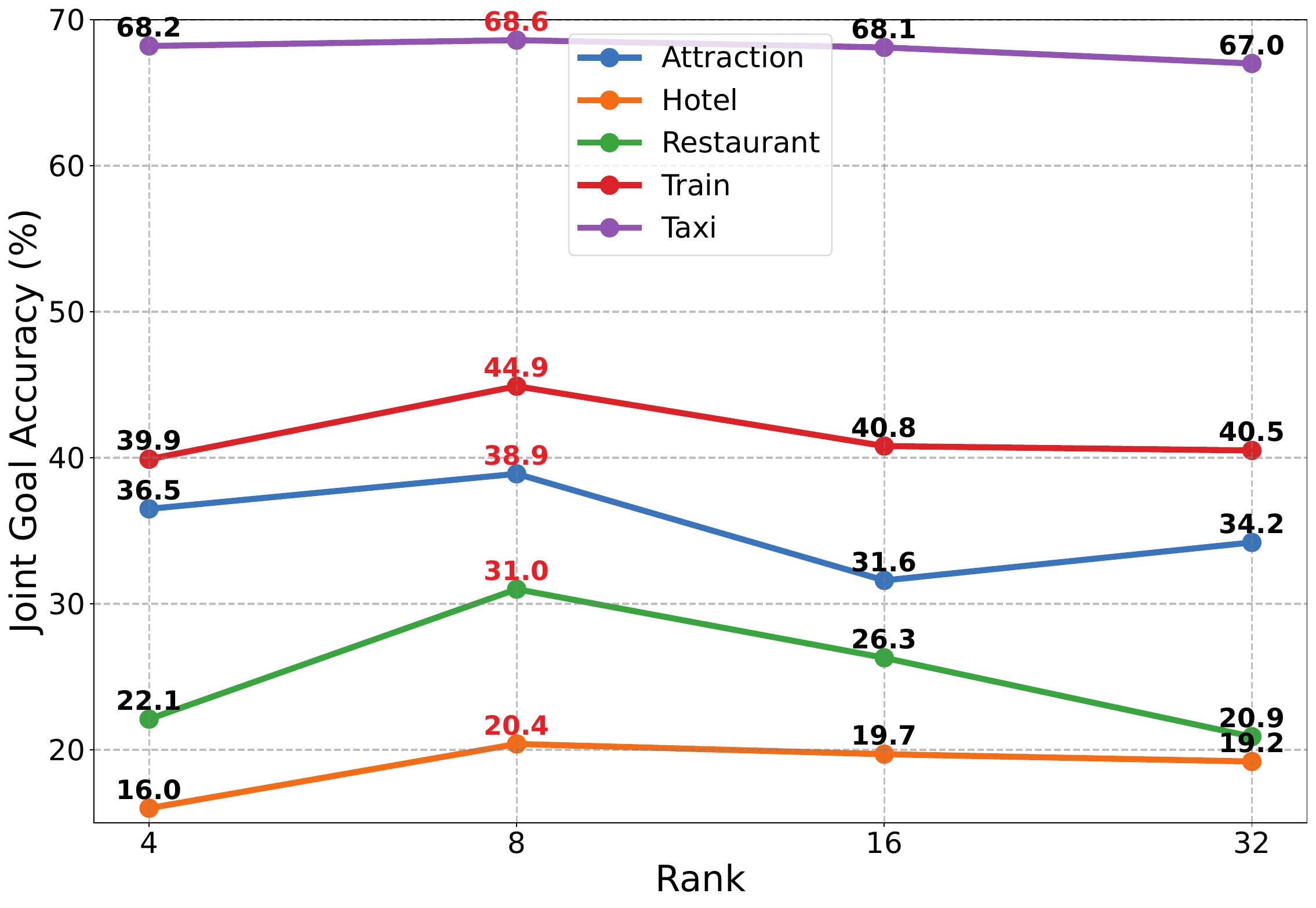}
        \caption{Accuracy of HiCoLoRA with different rank on the MultiWOZ dataset.}
        \label{fig: rank}
    \end{minipage}
    \hfill
    \begin{minipage}[t]{0.48\textwidth}
        \centering
        \includegraphics[width=\linewidth, keepaspectratio]{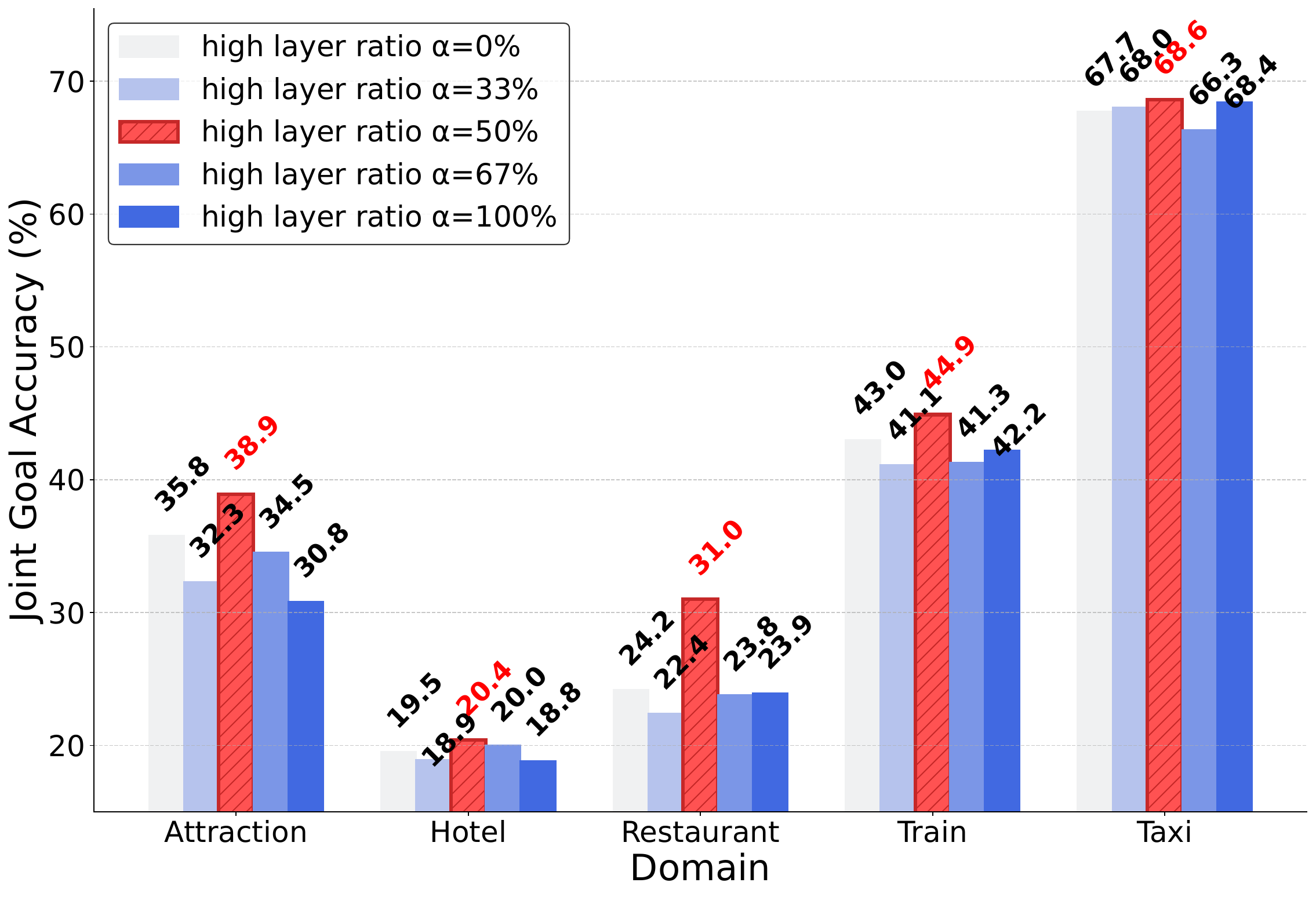}
        \caption{Accuracy of different high layer ratio (full collaboration) in HiCoLoRA.}
        \label{fig: ratio}
    \end{minipage}
\end{figure}

\noindent \textbf{Rank Sensitivity: Balance of expressiveness.}
Fig. \ref{fig: rank} shows that the superiority of $rank=8$ reflects LoRA principles: the rank must match the semantic complexity. Too low ($4$) fails to encode nuanced domain slot distinctions. Too high ($16/32$) introduces redundancy and dilutes transferable signals. This aligns with low-rank matrix theory, where rank determines perturbation precision to pre-trained weights, optimizing zero-shot transfer by balancing parsimony and expressiveness.


\noindent \textbf{High-Layer Ratio: Optimizing Semantic Flow.}
Fig. \ref{fig: ratio} indicates that the 50\% high-layer ratio validates cognitive theories of dialog comprehension, requiring balanced local-global integration. The 0\% ratio ignores global intent; 100\% dilutes slot-specific cues. HiCoLoRA's hierarchical design mirrors bottom-up (local atoms) to top-down (global intent) processing, ensuring coherent semantic chains, critical to resolving dynamic context-prompt misalignment in zs-DST.

\begin{figure}[t]
    \centering
    \begin{subfigure}{0.22\textwidth}
        \centering
        \includegraphics[width=\linewidth]{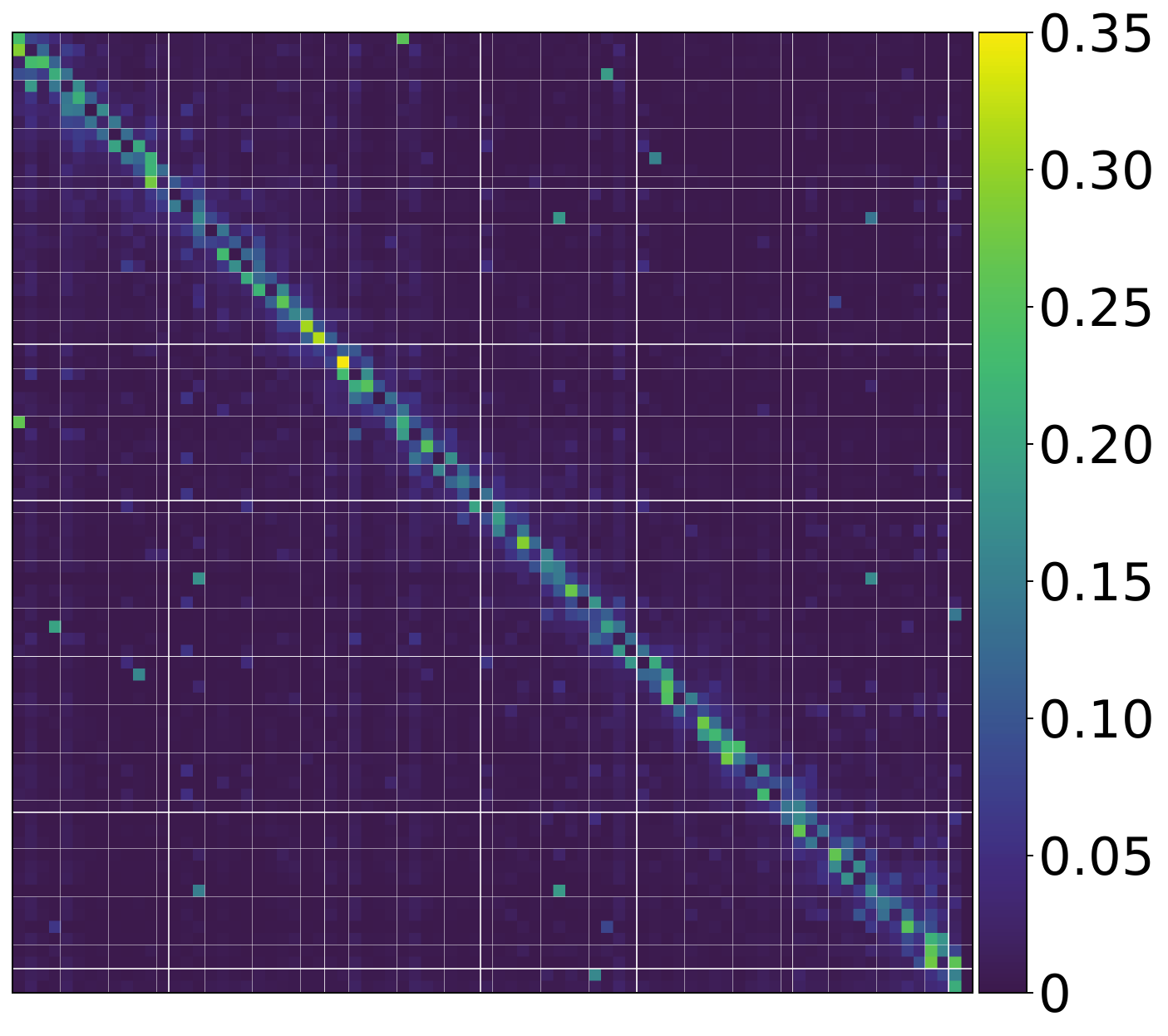} 
        \caption{First Layer Attention Map}
    \end{subfigure}
    \hfill
    \begin{subfigure}{0.22\textwidth}
        \centering
        \includegraphics[width=\linewidth]{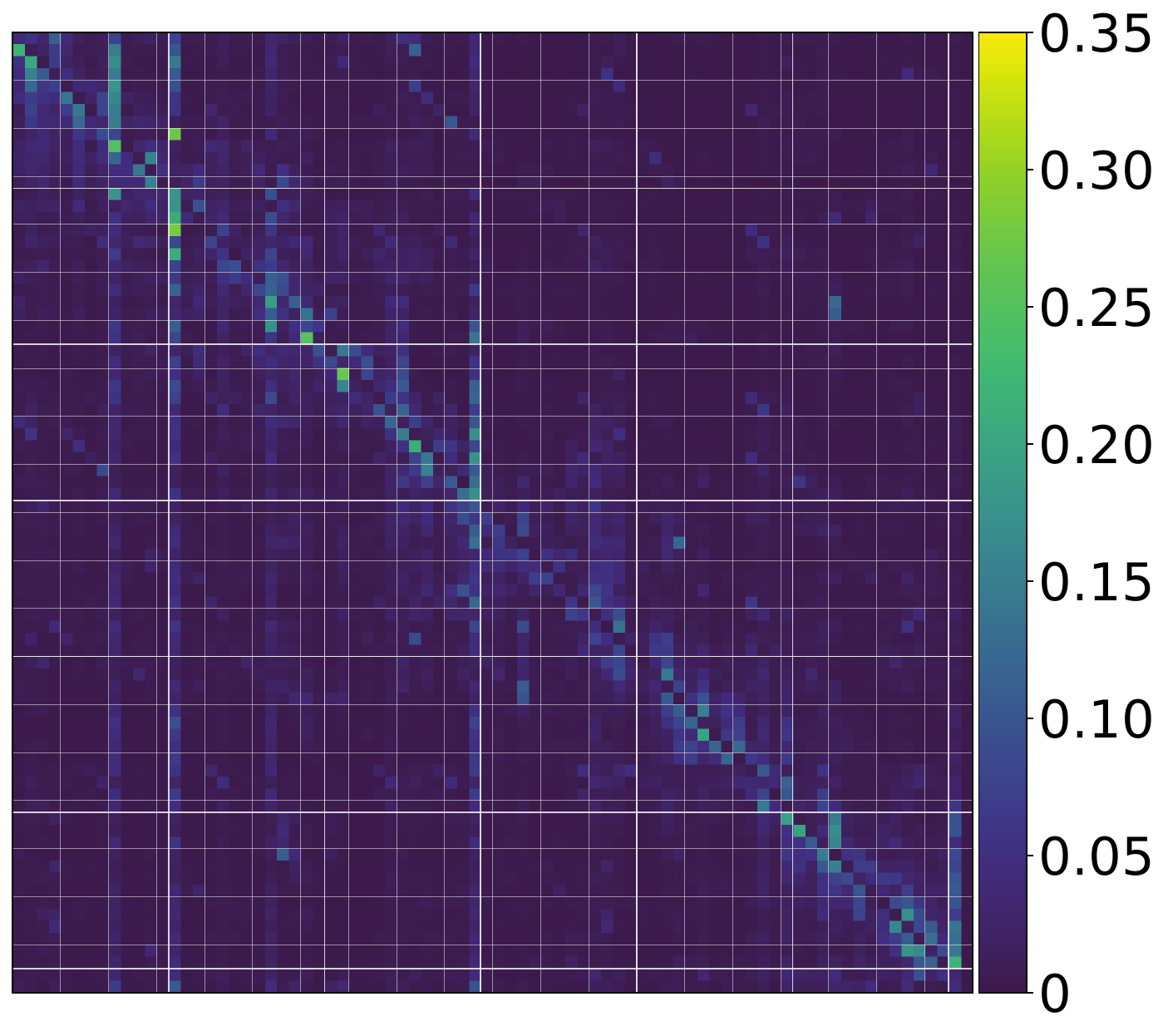} 
        \caption{Last Layer Attention Map}
    \end{subfigure}

    \caption{Example Attention Maps of the First and Last Transformer Layers in HiCoLoRA.}
    \label{fig: attn}
\end{figure}

\noindent \textbf{Attention Alignment: Maintaining Semantic Focus.}
Fig. \ref{fig: attn} reveals hierarchical attention evolution: first-layer ``local dots'' encode discrete context-prompt associations, while last-layer ``connected lines'' form global semantic chains. This mirrors the layered semantic progression of Transformer: lower layers anchor atomic prompt-semantic links, and higher layers integrate into coherent intent pathways through cross layer optimization. By preserving prompt focus across depths, HiCoLoRA avoids deep-layer attention dilution, maintaining critical alignment for zero-shot transfer.

The experimental results here validate our claims: optimal rank 8 confirms balanced expressiveness, the 50\% high-layer ratio verifies the optimization of semantic flow, and attention evolution demonstrates effective hierarchical collaboration. These align with the HiCoLoRA design, proving that its components jointly resolve misalignment.

\subsection{Case Study and Failure Analysis}
Our case study analysis in Appendix~\ref{sec:case_study} reveals HiCoLoRA's strengths in handling complex multi-domain dialogs through hierarchical collaboration and semantic disentanglement. Successful cases demonstrate robust slot inference in both transfer rich and context sensitive domains. However, failure patterns highlight areas for future refinement, particularly in highly idiosyncratic domains.

\subsection{Extended and Scalability Analysis}

\noindent \textbf{Scalability Analysis.}
HiCoLoRA exhibits enhanced scalability in larger datasets: 9.4\% average JGA gain in SGD vs 5.4\% in MultiWOZ. This stems from: 1) Semantic regular domains like \textit{Media} benefit from spectral clustering's cross domain pattern recognition; 2) Terminology intensive domains such as \textit{Flights} leverage SemSVD-Init's knowledge preservation; 3) Sparsely distributed slots like \textit{hotel-star} benefit from hierarchical refinement and singular value modulation.

\noindent \textbf{Extended Comparative Analysis.}
We conduct extensive comparisons against both recent LoRA methods and larger LLMs based approaches. As detailed in Appendix~\ref{sec:more_lora} -~\ref{sec:comparison-fnctod}, HiCoLoRA consistently outperforms recent LoRA variants in nearly all domains, achieving the highest average JGA. This superiority underscores the effectiveness of our hierarchical adaptation and semantic aware initialization in mitigating cross layer misalignment and knowledge distortion. Furthermore, when scaled to larger backbone models, HiCoLoRA remains highly competitive with other LLM-based zs-DST methods, even surpassing the previous SOTA FnCTOD, demonstrating its generalizability across model scales. These results confirm HiCoLoRA offers a robust and scalable solution for zero-shot dialog state tracking, effectively balancing performance and parameter efficiency.

\noindent \textbf{Architectural Implications beyond Homogeneous Baselines}  
The comparative analysis with heterogeneous methods reveals distinctive advantages of HiCoLoRA's design philosophy. While LDST relies on full fine-tuning of LLMs and CAPID introduces additional complexity through separate prompt generation, our approach demonstrates that a unified hierarchical architecture with collaborative adapters suffices to achieve competitive performance. This underscores the significance of structural alignment with task hierarchies over merely scaling model capacity or pipeline complexity, positioning HiCoLoRA as a resource efficient yet powerful paradigm for dialog state tracking.

\noindent \textbf{Generalization Analysis.} As analyzed in Appendix~\ref{sec:generalization}, our model demonstrates generalization capabilities in cross domain adaptation and long tailed recognition scenarios. It achieves performance improvements on multiple datasets, underscoring its ability to transfer knowledge across diverse domains. In addition, it exhibits remarkable robustness in tail classes, effectively mitigating the performance disparity between head and tail categories. This is attributed to our framework's ability to learn a more balanced and generalizable feature representation, which prevents overfitting to the dominant head classes and fosters a more robust decision boundary for underrepresented tail classes, thereby enhancing overall model generalization in real world and long tailed environments.
HiCoLoRA outperforms standard LoRA and DualLoRA on cross-task zero-shot transfer (Appendix~\ref{app:cross_task_transfer}), confirming that its semantic decoupling mechanism generalizes effectively to unseen slot distributions.

\noindent\textbf{Efficiency Analysis.}
Despite its hierarchical multi-branch design, HiCoLoRA maintains inference efficiency comparable to standard LoRA through a pre-computation and weight-merging strategy. By aggregating the low-rank matrices of SemAdapt-LoRA into a single pair \((\boldsymbol{A}_{\text{total}}, \boldsymbol{B}_{\text{total}})\) during inference (Appdendix~\ref{subsec:inference_efficiency}), the model incurs negligible overhead over single-adapter baselines. Benchmark results (Table~\ref{tab:inference_efficiency}) confirm that HiCoLoRA adds only minimal latency while delivering substantial performance gains, thus remaining practical for real-time dialog systems.

The extended analyses collectively affirm that HiCoLoRA's hierarchical adaptation transcends mere parameter efficiency by fundamentally restructuring semantic flow dynamics across transformer layers. Its spectral disentanglement mechanism effectively decouples domain agnostic and domain specific semantics, enabling robust knowledge transfer even under significant distribution shifts. This architectural paradigm demonstrates that task aligned inductive biases, rather than sheer model scale or pipeline complexity, constitute the pivotal factor for achieving scalable zero-shot generalization in dynamic dialog environments.

\section{Conclusion}
zs-DST is crucial for scalable TODs but remains challenged by insufficient cross-layer coordination, semantic conflation across domains, and corruption of pre-trained knowledge. HiCoLoRA overcomes these issues via a hierarchical LoRA design for dynamic context-prompt alignment, spectral clustering for domain-slot disentanglement, and SemSVD-Init for knowledge-preserving fine-tuning. Evaluations in MultiWOZ and SGD show that HiCoLoRA significantly outperforms previous SOTA approaches, improving average JGA by 5.4\% and 9.4\%, respectively. Limitations remain in highly idiosyncratic slot domains, and future work will focus on slot aware refinement to further strengthen HiCoLoRA's applicability in zs-DST.

\section*{Limitations}

While HiCoLoRA advances zs-DST performance, several limitations persist: the model struggles with ambiguous slot boundaries, leading to prediction errors when slot values overlap or are implicitly referenced; cross domain confusion arises in multi-domain dialogs where similar slot names cause semantic entanglement; rare or unseen slot values are poorly generalized, as the current initialization and adaptation mechanisms do not fully address domain specific sparsity; and highly idiosyncratic slots with domain-exclusive terms remain challenging, as spectral clustering may fail to capture low frequency semantic associations and higher layer fusion can dilute relevant signals. Future work may refine slot boundary detection, enhance domain disambiguation, and develop more targeted rare slot handling strategies.

\section*{Ethics Statement}
Our research involves only publicly available, anonymized dialog datasets (MultiWOZ and SGD) and does not collect new human subject data. Their use complies with the consent agreements established during their original release. All data usage complies with the original licenses, and no personally identifiable information is processed or stored. The proposed method, HiCoLoRA, is designed to improve zero-shot generalization in task-oriented dialog systems and does not have known harmful applications. While the method itself poses no direct ethical hazards, we note that any dialogue system carries a risk of misunderstanding user inputs or propagating biases if deployed without proper safeguards in sensitive applications. We acknowledge that there are no conflicts of interest and that the research was conducted with full integrity, transparency, and respect for privacy, fairness, and inclusivity. No institutional review board (IRB) approval was required as the study involves no human participants beyond the use of existing, de-identified benchmark data.

We acknowledge the use of Writeful integrated with Overleaf for refining the textual expression of this manuscript, and DeepSeek V3.2 for error correction of the experimental code. The role of these LLMs was limited to technical assistance and did not involve research ideation or the creation of core content. All LLM outputs have been rigorously verified by the authors, who bear full responsibility for the final accuracy, integrity, and originality of the content including the avoidance of plagiarism or scientific misconduct.

\section*{Acknowledgments}
This research is supported in part by the National Natural Science Foundation of China (No. 62472277), the National Natural Science Foundation of China (Youth) under Grant No. 62306208, the Shanghai East Talents Program (2023-177), the Shenzhen Medical Research Fund (No. D2404001), the Key Research and Development Program of Guangdong Province (No. 2025B1111020001), the Shenzhen Municipal STIB Key programs (No. CJGJZD20230724093303007 and KJZD20240903101259001), the National Key Laboratory of the CAS on Medical Imaging Science and Technology System, the Xisike Clinical Oncology Research Foundation (Y-2024AZ(NSCLC)MS-0156), and the SIAT-WUXI Joint Innov-Group for AGI-MET.


\bibliography{custom}

\appendix

\section{Experimental Details}
\label{sec:exp_details}

\subsection{Dataset Statistic}
\label{sec:dataset_stats}

\begin{table}[t]
\centering
    \centering
    {\fontsize{9}{10}\selectfont
    \begin{tabular}{@{}cccc@{}}
    \toprule
    \textbf{Domain} & Train & Dev  & Test \\ \midrule
    Attraction      & 2717  & 401  & 416  \\
    Hotel           & 3381  & 416  & 394  \\
    Restaurant      & 3813  & 438  & 207  \\
    Taxi            & 1654  & 207  & 195  \\
    Train           & 3103  & 484  & 494  \\ \midrule
    Total           & 8438  & 1000 & 1000 \\ \bottomrule
    \end{tabular}
    }
    \caption{The dataset statistic of MultiWOZ.}
    \label{tab: MultiWOZ statistic}
\end{table}

\begin{table}[t]
    \centering
    {\fontsize{9}{10}\selectfont
    \begin{tabular}{@{}cccc@{}}
    \toprule
    \textbf{Domain} & Train & Dev  & Test \\ \midrule
    Buses & 2,280 & 329 & 526 \\
    Events & 3,509 & 418 & 592 \\
    Flights & 2,747 & 391 & 506 \\
    Media & 1,113 & 179 & 364 \\
    Messaging & NA & NA & 298 \\
    Music & 1,290 & 196 & 347 \\
    Payment & NA & NA & 222 \\
    Trains & NA & NA & 350 \\ \midrule
    Total & 10,939 & 1,513 & 3,205 \\ \bottomrule
    \end{tabular}
    }
    \caption{The dataset statistic of SGD.}
    \label{tab: SGD statistic}
\end{table}

Based on the experimental design for zero-shot dialog state tracking, domain selection was strategically constrained to ensure robust evaluation. For MultiWOZ 2.1~\citep{eric-etal-2020-multiwoz} (Table \ref{tab: MultiWOZ statistic}), the Police (46 dialogs) and Hospital (38 dialogs) domains were excluded due to insufficient dialog volume and slot diversity, which would compromise statistical reliability in zero-shot generalization tests. Similarly, in SGD~\citep{Rastogi_Zang_Sunkara_Gupta_Khaitan_2020} (Table \ref{tab: SGD statistic}), services with limited samples or atypical slot structures, such as RideSharing (Test: 112), Calendar (Test: 98), etc., are omitted to avoid skew results. This curation focuses on evaluation on domains with adequate data density and representative slot semantics, ensuring that performance metrics reflect true zero-shot transferability rather than data-sparsity artifacts. Consequently, while coverage is reduced, the core challenge of cross-domain adaptation is preserved, with results generalizable to mainstream service-oriented interactions.

\subsection{Baseline Models}
\label{sec: baseline}

In this section, we provide a detailed overview of each baseline, as outlined below.

\subsubsection{Main Baseline}

\begin{itemize}
    \item \noindent {\bf TRADE} \cite{wu2019transferable} enhances dialog state generation by incorporating a copy mechanism and enabling knowledge transfer between tasks, allowing the model to handle unseen dialog states during training.

    \item \noindent {\bf MA-DST} \cite{Kumar2020MA} leverages cross-attention to align context and slot representations across multiple semantic levels, while using self-attention on RNN hidden states to resolve cross-domain coreference.

    \item \noindent {\bf SUMBT} \cite{lee2019sumbt}, built on the BERT-base, employs contextual semantic attention to learn the domain-slot-type and slot value relations, predicting slot values in a non-parametric manner.

    \item \noindent {\bf SGD-baseline} \cite{rastogi_scaling_2019} encodes dialog history and schema elements using BERT and applies conditional prediction with schema embeddings to accommodate dynamic schema sets.

    \item \noindent {\bf Seq2Seq-DU} \cite{feng_sequence--sequence_2021} formulates DST as a sequence-to-sequence task, using two BERT-based encoders to separately process dialog utterances and schema descriptions, followed by a pointer-based decoder to generate the dialog state.

    \item \noindent {\bf GPT2-DST} \cite{li-etal-2021-zero} utilizes a GPT2-base generative question answering model, enabling natural language queries to infer unseen constraints and slots for zero-shot generalization in multi-domain task-oriented dialogs.

    \item \noindent {\bf TransferQA} \cite{lin_zero-shot_2021} integrates extractive and multiple-choice question answering within a unified text-to-text transformer framework, effectively tracking both categorical and non-categorical slots, and introducing unanswerable questions to improve robustness.

    \item \noindent {\bf T5DST} \cite{lin_leveraging_2021}, based on T5-small and PPTOD-small, encodes dialog context and slot descriptions and generates slot values in an autoregressive manner. Slot-type descriptions facilitate cross-slot information sharing and cross-domain knowledge transfer.

    \item \noindent {\bf SlotDM-DST} \cite{wang-etal-2022-slot}, leveraging T5-small, models slot-slot, slot-value, and slot-context dependencies via slot prompts, value demonstrations, and constraint objects. Shared prompts capture transferable knowledge across domains.
    
    \item \noindent {\bf Prompter} \cite{aksu2023prompter}, based on PPTOD-small, generates dynamic prefixes from slot descriptions and injects them into the key and value states of each Transformer layer's self-attention mechanism, enabling zero-shot prefix tuning.
    
    \item \noindent {\bf DCC} \cite{wang_divide_2023} Divide, Conquer and Combine, built on T5-small, adopts a mixture-of-experts strategy by partitioning semantically independent data subsets, training corresponding experts, and applying ensemble inference for unseen samples.

    \item \noindent {\bf DualLoRA} \cite{ luo2024zero} builds on PPTOD-small with a T5-small backbone, employing two low-rank adaptation matrices, one refining dialog context and the other slot prompts.  Once trained, these matrices are fused into the frozen pre-trained weights, yielding zero-shot cross-domain dialog state tracking without any extra inference latency.

\end{itemize}

\subsubsection{LoRA Baseline}

\begin{itemize}
\item \noindent {\bf HydraLoRA} \cite{tian2024hydralora} is a parameter-efficient fine-tuning (PEFT) framework designed to address the performance gap between standard LoRA and full fine-tuning, especially on complex datasets. Introduce an asymmetric LoRA structure that does not require domain expertise. Experiments demonstrate that HydraLoRA surpasses existing PEFT methods in performance.

\item \noindent {\bf LoRA-GA}\ \cite{wang2024lora} improves LoRA by proposing a novel gradient-aware initialization strategy that aligns the gradients of the low-rank matrices with those of full fine-tuning at the first training step. This method significantly accelerates convergence (2-4× faster than vanilla LoRA) and improves performance in tasks such as GLUE, GSM8K, and code generation, even for large models such as Llama 2-7B.

\item \noindent {\bf RoSA} \cite{nikdan2024rosa}, Robust Adaptation combines low-rank and sparse adaptations inspired by robust PCA to approximate full fine-tuning performance under constrained computational budgets. It is particularly effective in generative tasks like math problem solving and SQL generation, and supports efficient training via custom sparse GPU kernels and compatibility with quantized base models.

\item \noindent {\bf Spectral Adapter} \cite{zhang2024spectral} incorporates spectral information from pretrained weights via SVD to enhance PEFT methods. Performs additive tuning or orthogonal rotation on the top singular vectors, improving rank capacity and parameter efficiency. The adapter also benefits multi-adapter fusion and demonstrates stronger performance across various tasks.

\end{itemize}

\subsubsection{LLM Baseline}

\begin{itemize}
\item \noindent {\bf ChatGPT-zsTOD} \cite{heck-etal-2023-chatgpt} achieves state-of-the-art performance in zero-shot dialog state tracking without task-specific training, leveraging its general-purpose language model capabilities. However, inherent limitations prevent it from fully replacing specialized systems, though its in-context learning abilities may support the development of dynamic dialog state trackers.

\item \noindent {\bf D0T} \cite{finch-choi-2024-diverse} enhances zero-shot DST by generating synthetic data across over 1,000 domains, creating a diverse training dataset with silver-standard annotations. This approach addresses data scarcity and enables adaptation to new domains without costly collection efforts.

\item \noindent {\bf MoPE} \cite{tang-etal-2024-mope} proposes a Mixture of Prefix Experts to connect similar slots across different domains, improving transfer performance in unseen domains. It addresses domain transferring and partial prediction problems in zero-shot DST.

\item \noindent {\bf FnCTOD} \cite{li2024large} improves zero-shot DST by calling functions with LLMs, allowing adaptation to diverse domains without extensive data or tuning. It achieves state-of-the-art performance with both open-source and proprietary LLMs, significantly boosting ChatGPT and GPT-4 results.

\item \noindent {\bf Multi-User} \cite{song2025multiuser} evaluates LLMs in multi-user DST by extending datasets with second-user utterances generated via speech act theory. For a fair comparison, the experimental setup was configured using single-user data to evaluate the performance of LLMs in single-user dialog state tracking.
\end{itemize}

\subsection{Evaluation Metric Formulas}
\label{sec:formula}
\subsubsection{JGA Formula} 
 \begin{equation} 
 \mathit{JGA} = \frac{\sum_{i=1}^{T} I( S_{i}^{pre} = S_{i}^{gt})}{T} 
 \label{equ:JGA} 
 \end{equation} 
 In this formula, $T$ denotes the total number of dialog turns in the evaluation dataset. For each turn $i$, $S_{i}^{pre}$ and $S_{i}^{gt}$ represent the predicted and ground truth sets of slot-value pairs, respectively. The indicator function $I$ returns 1 if the inside condition is satisfied and 0 otherwise. Specifically, $I(S_{i}^{pre} = S_{i}^{gt})$ checks whether the predicted set of slot-value pairs for turn $i$ exactly matches the set of ground truth slot-value pairs. A value of 1 indicates a perfect match for that turn, that is, all slot value pairs were correctly predicted, while any discrepancy results in a value of 0. The summation $\sum_{i=1}^{T} I(S_{i}^{pre} = S_{i}^{gt})$ thus counts the number of turns for which the entire set of slot-value pairs was correctly predicted. 

\subsubsection{AGA Formula}
\begin{equation}
\mathit{AGA} =
\frac{\sum_{i=1}^{T} \frac{|S_{i}^{gt} \cap S_{i}^{pre}| - |S_{i}^{pre} - S_{i}^{gt}| {unique}}{|S {i}^{gt}|}}{T}
\label{equ:AGA}
\end{equation}
In this formula, $T$ denotes the total number of dialog turns in the evaluation dataset. For each turn $i$, $S_{i}^{pre}$ and $S_{i}^{gt}$ represent the predicted and ground truth sets of slot-value pairs, respectively. The formula calculates the slot-level accuracy for each turn by:
\begin{itemize}
\item Computing the intersection $|S_{i}^{gt} \cap S_{i}^{pre}|$, which counts correctly predicted slot-value pairs
\item Computing $|S_{i}^{pre} - S_{i}^{gt}| {unique}$, which counts incorrectly predicted slots (by extracting unique slot names from the difference set)
\item Subtracting incorrect predictions from correct predictions
\item Normalization by the total number of ground truth slot-value pairs $|S {i}^{gt}|$
\end{itemize}
The outer summation averages these per-turn accuracies across all dialog turns. Note that this is a more complex metric than simple slot matching, as it accounts for both missed slots and incorrect slot predictions while considering slot name uniqueness.

\subsection{Experiments Implementation Details}
\label{sec:exp_impl_details}
Our experimental setup, designed for a precise comparison with previous work, follows that of DualLoRA \cite{luo2024zero}. We use the T5-small architecture (6 encoder/decoder layers, 512 hidden dimension, 8 attention heads) as the backbone for HiCoLoRA, with a LoRA rank of 8 for low-rank adaptation, initialized from PPTOD-small checkpoints, consistent with observations in DualLoRA that PPTOD \cite{su_multi-task_2022} is particularly suitable for prompt-tuning due to its pre-training objectives.

For spectral clustering, the number of domain clusters ($M$) and slot clusters ($N$) are set as 2 and 3 for MultiWOZ, with 2 and 4 specified for SGD. These configurations are determined by maximizing the silhouette coefficient. These configurations are determined by maximizing the silhouette coefficient using only source domain schemas, ensuring strict zero-shot compliance (no target domain information is used).

Training configurations include a batch size of 8 with gradient accumulation every 8 steps, the AdamW optimizer (weight decay 0.01, learning rate 1e-4, no scheduler), a fixed random seed of 3407 for ablation studies, and 5 training epochs (early stopping after 5 consecutive validation loss plateaus). For main results, we report mean $\pm$ standard deviation over five random seeds: \{3407, 42, 100, 2345, 5050\}.

For hierarchical processing, we use a $\alpha=50\%$ full collaboration ratio with higher layers  and a semantic enhancement coefficient $\lambda=0.5$ to modulate singular values in semantically enhanced SVD initialization.

The training and validation sets exclude target domain data, while the test set retains only target domain instances. All experiments were conducted on NVIDIA GeForce RTX 5080 GPU and Python 3.10.

\subsection{End-to-End Pipeline and Domain Separation}
\label{sec:pipeline}

\textbf{Formal Domain Separation.} 
To ensure strict zero-shot evaluation, we partition all domains $\mathcal{D}_{\text{all}}$ into disjoint training (source) and testing (target) sets: 
$\mathcal{D}_{\text{seen}} \cap \mathcal{D}_{\text{unseen}} = \emptyset$. 
The training and validation sets exclude any dialogue containing a target domain, while the test set retains only target-domain instances.

\textbf{Training Pipeline.} 
The end-to-end training consists of four sequential stages:
\begin{enumerate}
    \item \textit{Spectral Clustering (offline):} Encode domain names and slot prompts via T5, then apply spectral clustering to obtain domain clusters $\mathcal{D}^M$ and slot clusters $\mathcal{X}^N$ (Sec.~3.4).
    \item \textit{Initialization:} Initialize LoRA matrices via SemSVD-Init using pre-trained weights (Sec.~3.5).
    \item \textit{Hierarchical Optimization:} Train UniRep-LoRA and SemAdapt-LoRA with layer-specific strategies (Sec.~3.2, 3.3): lower layers use heuristic grouping, higher layers use full collaboration.
    \item \textit{Adaptive Fusion:} Dynamically balance domain-agnostic and domain-specific representations via learnable gating $\beta$ (Eq.~4).
\end{enumerate}

\textbf{Inference Pipeline.} 
Despite multi-branch design, we precompute aggregated matrices during model export:
$\boldsymbol{A}_{\text{total}} = \sum_{m=1}^M \boldsymbol{A}_{sa}^m$, $\boldsymbol{B}_{\text{total}} = \sum_{n=1}^N \boldsymbol{B}_{sa}^n$.
The forward pass then reduces to a single low-rank update:
$\boldsymbol{h}_{sa} = \boldsymbol{W}_0^h \boldsymbol{x}_{sa} + \boldsymbol{B}_{\text{total}} \boldsymbol{A}_{\text{total}} \boldsymbol{x}_{sa}$.
This eliminates online overhead and maintains latency comparable to standard LoRA (Table~\ref{tab:inference_efficiency}).

\section{Additional Results and Analyses}
\label{sec: more_table}
\subsection{Performance on SGD Dataset}
\label{sec:sgd_result}

\begin{table*}[t]
\centering
{\fontsize{9}{11}\selectfont
\resizebox{\textwidth}{!}{
\begin{tabular}{@{}lccccccccc@{}}
\toprule
\textbf{Method} & \textbf{Year} & \textbf{Buses} & \textbf{Events} & \textbf{Flights} & \textbf{Media} & \textbf{Messaging} & \textbf{Music} & \textbf{Payment} & \textbf{Trains} \\ 
\midrule
SGD-baseline & 2019 & 9.7/50.9 & 23.5/57.9 & 23.9/65.9 & 18.0/30.8 & 10.2/20.0 & 15.5/39.9 & 11.5/34.8 & 13.6/63.5 \\
Seq2seq-DU & 2021 & 16.8/N & 31.9/N & 15.9/N & 23.1/N & 4.9/N & 12.3/N & 7.2/N & 16.8/N \\
Transfer-QA & 2021 & 15.9/63.6 & 15.6/56.8 & 3.59/42.9 & 30.2/67.5 & 13.3/37.9 & 8.9/62.4 & 24.7/60.7 & 17.4/64.9 \\
SlotDM-DST & 2022 & 43.9/86.3 & \text{--} & \text{--} & \text{--} & 36.6/61.4 & \text{--} & 16.5/62.0 & 46.7/86.9 \\
T5DST & 2021 & 46.8/N & 48.8/N & \text{--} & 55.5/N & 59.2/N & \text{--} & 23.3/N & 53.0/N \\
Prompter & 2023 & 48.4/N & 51.5/N & \text{--} & 65.3/N & 59.2/N & \text{--} & 21.9/N & 50.8/N \\
DCC & 2023 & \text{--} & \text{--} & \text{--} & \text{--} & 28.8/N & \text{--} & 19.4/N & 42.3/N \\
DualLoRA (Prev. SOTA) & 2024 & 50.9/88.8 & 46.5/82.8 & 28.4/76.9 & 69.7/88.7 & 65.1/85.5 & 32.5/72.4 & 21.2/\textbf{70.2} & 52.9/89.3 \\
\cmidrule(lr){1-10}
HiCoLoRA (Ours) & 2025 & 
\shortstack{54.17$\pm$0.28 \\ /93.2} &
\shortstack{55.01$\pm$0.14 \\ /87.8} &
\shortstack{30.57$\pm$0.12 \\ /82.3} &
\shortstack{76.20$\pm$0.21 \\ /95.8} &
\shortstack{67.79$\pm$0.22 \\ /88.1} &
\shortstack{35.46$\pm$0.30 \\ /78.9} &
\shortstack{26.98$\pm$0.21 \\ /65.0} &
\shortstack{55.99$\pm$0.21 \\ /93.8} \\
\cmidrule(lr){1-10}
\textbf{\% Gain vs DualLoRA} & - & \textbf{+6.1/+5.0} & \textbf{+18.5/+6.0} & \textbf{+8.1/+7.0} & \textbf{+8.9/+8.0} & \textbf{+4.0/+3.0} & \textbf{+10.2/+9.0} & \textbf{+25.9/-7.4} & \textbf{+5.5/+5.0} \\
\bottomrule
\end{tabular}
}
}
\caption{Zero-shot JGA (\%) \& AGA (\%) on the SGD dataset with relative improvements over previous SOTA. ``N'' indicates unreported results.}
\label{table: SGD_modified}
\end{table*}
Table~\ref{table: SGD_modified} presents the zero-shot performance of HiCoLoRA on the SGD Dataset. Compared to baseline methods and previous state-of-the-art approaches, HiCoLoRA achieves significant improvements across multiple domains.

\subsection{Comparison with Contemporary LoRA Methods}
\label{sec:more_lora}
\begin{table*}[t]
\centering
{\fontsize{9}{11}\selectfont
\setlength{\tabcolsep}{1mm}
\begin{tabular}{@{}lccccccc@{}}
\toprule
\textbf{Method} & \textbf{Year} & \textbf{Attr.} & \textbf{Hotel} & \textbf{Rest.} & \textbf{Train} & \textbf{Taxi} &  \textbf{AVG.} \\
\midrule
HydraLoRA & 2024 & 35.1 & 18.9 & 26.3 & 41.5 & 65.2 & 37.4 \\
LoRA-GA & 2024 & 33.8 & 19.2 & 24.7 & 42.8 & 64.1 & 36.9 \\
RoSA & 2024 & 36.5 & 19.6 & 27.9 & 43.2 & 66.8 & 38.8 \\
Spectral Adapter & 2025 & 37.2 & 20.1 & 28.5 & 43.6 & 67.3 & 39.3 \\
HiCoLoRA (Ours) & 2025 & \textbf{38.9} & \textbf{20.4} & \textbf{31.0} & \textbf{44.9} & \textbf{68.6} & \textbf{40.8} \\
\bottomrule
\end{tabular}
}
\caption{Comparison of HiCoLoRA with recent LoRA-based methods on MultiWOZ (JGA \%).}
\label{table: more_lora}
\end{table*}

To situate HiCoLoRA within the evolving landscape of PEFT methods, we compare it against four contemporary LoRA variants: HydraLoRA \cite{tian2024hydralora}, LoRA-GA \cite{wang2024lora}, RoSA \cite{nikdan2024rosa}, and Spectral Adapter \cite{zhang2024spectral}. As shown in Table \ref{table: more_lora}, HiCoLoRA achieves the highest average JGA, outperforming all baselines in nearly all domains. This superiority is not merely incremental; it stems from fundamental architectural and semantic distinctions that address the core challenges of zs-DST.

\textbf{Structural Design Philosophy:}
While HydraLoRA introduces an asymmetric LoRA structure to enhance expressiveness, and RoSA combines low-rank and sparse adaptations for robustness, both methods retain a \textit{layer-agnostic} approach to adapter deployment. In contrast, HiCoLoRA's \textit{hierarchical layer-specific processing} explicitly models the divergent roles of lower and higher Transformer layers, local feature encoding versus global intent integration, enabling dynamic cross-layer coordination that is critical for resolving context-prompt misalignment.

\textbf{Semantic Alignment Mechanism:}
Spectral Adapter leverages spectral initialization to better preserve pre-trained knowledge, similar to our SemSVD-Init. However, it lacks HiCoLoRA's \textit{spectral joint clustering} of domains and slots, which actively disentangles domain-shared and domain-specific semantics. This clustering guides the adaptive fusion of general and domain-aware features, a mechanism absent in other methods, leading to more precise slot inference in transfer-rich domains like \textit{Media}.

\textbf{Knowledge Preservation and Transfer:}
LoRA-GA improves the alignment of the gradient during initialization to accelerate convergence but does not explicitly modulate the singular values to align with the specific semantics of the task. HiCoLoRA's \text{SemSVD-Init} not only preserves pre-trained knowledge, but also amplifies singular components relevant to domain-slot structures, effectively mitigating catastrophic forgetting and enhancing zero-shot generalization, particularly for rare slots such as \textit{hotel-stars}.

\textbf{Adaptability to Dynamic Contexts:}
Unlike RoSA and HydraLoRA, which are designed for general NLP tasks, HiCoLoRA is tailored for the dynamic and multi-turn nature of dialog systems. Its \textit{adaptive gating mechanism} dynamically balances domain-agnostic and domain-specific features per turn, enabling robust handling of evolving dialog contexts, a capability that static LoRA variants lack.

HiCoLoRA addresses the unique challenges of zs-DST: cross-layer misalignment, semantic conflation, and knowledge distortion. While other LoRA variants offer general-purpose efficiency, HiCoLoRA provides a \textit{domain-aware} and \textit{layer-conscious} design that is essential for robust zero-shot transfer in TODs.

\subsection{Scalability Analysis: Generalization Across Model Scales}
\label{sec:more_llm}
\begin{table*}[t]
\centering
{\fontsize{9}{11}\selectfont
\setlength{\tabcolsep}{1mm}
\begin{tabular}{@{}lccccccccc@{}}
\toprule
\textbf{Method} & \textbf{Year} & \textbf{Base Model} & \textbf{Attr.} & \textbf{Hotel} & \textbf{Rest.} & \textbf{Train} & \textbf{Taxi} &  \textbf{AVG.} \\
\midrule
ChatGPT-zsTOD & 2023 & ChatGPT (GPT-3.5) & 52.7 & 42.0 & 60.8 & 70.9 & 55.8 & 56.4 \\
ChatGPT-zsTOD & 2023 & ChatGPT (GPT-3.5) & 67.2 & 37.6 & 67.3 & \textbf{74.4} & 60.1 & 61.3 \\
D0T & 2024 & LLAMA2-13B & 63.1 & 43.8 & 60.8 & 48.8 & 64.7 & 56.2 \\
MoPE & 2024 & ChatGLM-6B & 60.4 & 34.1 & \textbf{64.0} & 71.3 & 55.9 & 57.1 \\
FnCTOD & 2024 & ChatGPT (GPT-4) & 58.8 & 45.2 & 69.5 & 76.4 & 63.2 & \textbf{62.6} \\
FnCTOD & 2024 & LLAMA2-13B & 62.2 & \textbf{46.8} & 60.9 & 67.5 & 60.3 & 59.5 \\
Multi-User & 2025 & GPT-4o & 56.8 & 46.0 & 61.9 & 69.3 & 55.1 & 57.8 \\
\midrule
HiCoLoRA & 2025 & LLAMA2-13B & 62.0 & 42.0 & 61.0 & 65.0 & 69.0 & 60.0 \\
HiCoLoRA & 2025 & Qwen2.5-14B-Instruct & \bf{64.0} & 44.0 & 63.0 & 68.0 & \textbf{71.0} & 62.0 \\
\bottomrule
\end{tabular}
}
\caption{Zero-shot JGA (\%) on MultiWOZ using large language models. HiCoLoRA demonstrates strong scalability and generalization across model scales. All results of baselines were reported from original papers.}
\label{table: more_llm}
\end{table*}

To rigorously assess the scalability and architectural generality of HiCoLoRA, we extend our evaluation to LLM, comparing against contemporary LLM-based zs-DST methods, including ChatGPT-zsTOD \cite{heck-etal-2023-chatgpt}, D0T \cite{finch-choi-2024-diverse}, MoPE \cite{tang-etal-2024-mope}, FnCTOD \cite{li2024large} and Multi-User \cite{song2025multiuser}. As shown in Table \ref{table: more_llm}, HiCoLoRA achieves competitive performance when deployed in LLAMA2-13B and Qwen2.5-14B-Instruct, with an average JGA of 62.0\% in the latter, only marginally below FnCTOD with GPT-4 (62.6\%) and significantly outperforms other baselines based on LLM.

\noindent\textbf{Architectural Generalization Beyond Scale.}
The consistent performance of HiCoLoRA in both both small (T5-small, 60M) and large (13B-14B) models underscores a key insight: its hierarchical adaptation mechanism is \textit{scale-agnostic}. The efficacy of HiCoLoRA stems from its structured semantic alignment decomposition, which addresses cross-layer coordination (RQ1), domain-slot disentanglement (RQ2), and knowledge preservation (RQ3) through explicit inductive biases. This allows it to be generalized effectively even when applied to larger models without architecture-specific modifications.

\noindent\textbf{Efficiency-Performance Trade-off.}
While FnCTOD benefit from extreme scale and extensive pre-training as GPT-4-based methods, HiCoLoRA offers a more efficient alternative, achieving comparable performance with only partial parameter updates. This highlights its suitability for scenarios where full fine-tuning or inference with very large models is prohibitive. The fact that HiCoLoRA outperforms other PEFT-based LLM methods further validates its superior design in leveraging limited tunable parameters for maximal semantic alignment.

\subsection{Extended Comparison with FnCTOD}
\label{sec:comparison-fnctod}

Since FnCTOD~\citep{li2024large} achieves comparable performance to HiCoLoRA under the same LLaMA2-13B backbone, we conduct a detailed comparison to highlight their differences in experimental setup and efficiency. A thorough examination of FnCTOD's experimental configuration reveals several deviations from a strict zero-shot setting. 

FnCTOD uses a carefully curated dataset of 7,200 dialogs across 36 domains (including SGD, CamRest676, MSR-E2E, TaskMaster, and WOZ), which include domains overlapping with MultiWOZ test domains. This violates the strict zero-shot learning premise. In contrast, HiCoLoRA uses only 4,625-7,684 samples from 4 domains in MultiWOZ, with one domain excluded during training to ensure a strict zero-shot setting. To ensure a fair comparison, we conducted an additional experiment by training FnCTOD on the \textbf{FnCTOD dataset}. The results, summarized in Table~\ref{tab:fnctod-comparison}, demonstrate that FnCTOD achieves superior performance while maintaining significantly higher efficiency.

\begin{table*}[t]
\centering
{\fontsize{9}{11}\selectfont
\resizebox{\textwidth}{!}{
\begin{tabular}{lccccccc}
\hline
\textbf{Method} & \textbf{Attr.} & \textbf{Hotel} & \textbf{Rest.} & \textbf{Train} & \textbf{Taxi} & \textbf{AVG.} & \textbf{Relative Change} \\
\hline
FnCTOD (Fine-tuned LLaMA2-13B) & 62.2 & 46.8 & 60.3 & 60.9 & 67.5 & 59.5 & -0.8 \\
FnCTOD (No FT LLaMA2-13B) & 49.8 & 29.5 & 48.9 & 53.6 & 64.7 & 49.3 & -21.1 \\
FnCTOD (GPT-4 SOTA) & 58.8 & 45.2 & 63.2 & 69.5 & \textbf{76.4} & 62.6 & +4.2 \\
HiCoLoRA (LLaMA2-13B) & 62.0 & 42.0 & 61.0 & 65.0 & 69.0 & 60.0 & -- \\
HiCoLoRA (FnCTOD Dataset) & \textbf{62.8} & \textbf{49.2} & \textbf{63.9} & \textbf{70.3} & 69.4 & \textbf{63.1} & +5.2 \\
\hline
\end{tabular}
}}
\caption{Performance comparison between HiCoLoRA and FnCTOD under different settings on MultiWOZ (JGA \%).}
\label{tab:fnctod-comparison}
\end{table*}

Beyond the fundamental discrepancy in training data composition, our comparative analysis reveals several critical distinctions that underscore HiCoLoRA's methodological rigor and practical efficiency: (1) When trained on identical data, HiCoLoRA achieves a JGA of 63.1, surpassing FnCTOD by 6.1\% and even exceeding GPT-4-based FnCTOD by 0.5 JGA points; (2) HiCoLoRA maintains superior inference efficiency, requiring only a single LLM call with 16 token prompts versus FnCTOD's dual invocations and larger than 1200 token inputs; (3) While FnCTOD(without fine-tune) employs 5 few-shot examples in its zero-shot configuration (achieving only 49.3 JGA), HiCoLoRA operates under strict zero-shot conditions to attain 60.0 JGA; (4) FnCTOD's incorporation of detailed schema descriptions deviates from minimal prompt principles, whereas HiCoLoRA relies solely on its hierarchical adaptation mechanism; (5) Architecturally, HiCoLoRA achieves competitive performance through semantic aware initialization and efficient parameter updates, avoiding the computational overhead of prompt heavy approaches.

This comparative analysis demonstrates that FnCTOD not only achieves state-of-the-art performance under strict zero-shot settings but also offers superior efficiency and scalability compared to prompt heavy LLM-based approaches. The gains are attributable to its principled hierarchical adaptation, spectral semantic disentanglement, and knowledge preserving initialization mechanisms that are both empirically effective and practically efficient.

\subsection{Ablation Study Results}
\label{sec: ablation_table}

\begin{table*}[ht]
\centering
{\fontsize{10}{10}\selectfont
\setlength{\tabcolsep}{1.3mm}
\begin{tabular}{@{}lccccccc@{}}
\toprule
\textbf{Method} & \textbf{Attr.} & \textbf{Hotel} & \textbf{Rest.} &\textbf{Train} & \textbf{Taxi} &  \textbf{AVG.} & \textbf{$\Delta$} \\ 
\midrule
HiCoLoRA (Full) & \textbf{38.9} & \textbf{20.4} & \textbf{31.0} & \textbf{44.9} & \textbf{68.6} & \textbf{40.8} & - \\
w/ Swap Hier Strategies & 37.2 & 19.7 & 22.9 & 40.2 & 67.5 & 37.4 & -3.4 \\
w/o Adaptive Fusion & 28.9 & 19.3 & 20.3 & 43.0 & 68.0 & 35.9 & -4.9 \\
w/o Spec Joint Cluster & 36.2 & 19.8 & 27.5 & 42.1 & 63.6 & 37.8 & -3.0 \\
w/ Kaiming Init & 34.3 & \textbf{20.4} & 27.8 & 40.4 & 67.5 & 38.1 & -2.7 \\
w/ PiSSA Init & 36.5 & 20.3 & 29.0 & 42.5 & 67.8 & 38.9 & -1.9 \\
w/ MiLoRA Init & 34.1 & 19.9 & 26.2 & 38.5 & 62.9 & 36.3 & -4.5 \\
\bottomrule
\end{tabular}
}
\caption{Ablation study on hierarchical architecture, adaptive fusion, spectral clustering, and initialization of HiCoLoRA on MultiWOZ. Attr. and Rest. are abbreviations for Attraction and Restaurant, respectively. The \textbf{$\Delta$} column shows the absolute performance drop compared to the full model.}
\label{table: ablation}
\end{table*}

Table~\ref{table: ablation} validates the contributions and necessity of each core component of HiCoLoRA to its overall performance. This validation is conducted by systematically removing or replacing core components, including the hierarchical strategy, adaptive fusion, spectral clustering, and initialization method.

\subsection{Comparison with Recent Heterogeneous Methods}
\label{sec:heterogeneous-comparison}

To further validate the effectiveness of HiCoLoRA against contemporary approaches with different architectural paradigms, we conducted comparative analyses with two recently proposed state-of-the-art methods: LDST~\citep{feng-etal-2023-towards} and CAPID~\citep{dong-etal-2024-zero}. 

\noindent \textbf{Comparison with LDST (EMNLP 2023):}
LDST proposes an Assembled Domain-Slot Instruction Generation approach for DST. This method generates diverse instruction samples by randomly combining different instruction and input templates during fine-tuning, thereby reducing the model's sensitivity to prompt variations. For example:


\begin{promptbox}
Instruction:
    Track the state of the slot <hotel-area> in the input dialog.
Input:
    [USER] I need to book a hotel in the east that has 4 stars. 
    [SYSTEM] I can help you with that. What is your price range? 
    [domain] hotel, [slot] area, it indicates area or place of the hotel. 
    This slot is categorical and you can only choose from the following available values: center, east, north, south, west. 
    If the slot is not mentioned in the dialog, just return NONE.
    So the value of slot <hotel-area> is
\end{promptbox}

We performed comparative experiments on MultiWOZ 2.1 using the LLaMA-7B backbone for both methods. The results demonstrate that HiCoLoRA maintains 1.9\% advantage over LDST (57.8 vs. 56.7 Average JGA). This performance gain, coupled with HiCoLoRA's parameter efficient design, further validates the effectiveness of our hierarchical collaborative architecture in capturing complex dialog state dependencies.

\noindent \textbf{Comparison with CAPID (EMNLP 2024):}
CAPID proposes Context-aware Auto-prompting and Instruction-following Contrastive Decoding. This approach employs a two stage framework where a context-aware slot query generation method via auto-prompting which initially using GPT-4, aligns the gap between source and target domains. The generated prompts are used to train a T5-base student model to independently produce context-aware slot queries. During inference, the fine-tuned T5-base student model first generates the prompt, which is then used by the trained DST model (T5-base or T5-small) to predict slot values.

\begin{table*}[t]
\centering
{\fontsize{9}{10}\selectfont 
\begin{tabular}{lccccccc}
\toprule
\textbf{Method} & \textbf{Configuration} & \textbf{Attr.} & \textbf{Hotel} & \textbf{Rest.} & \textbf{Train} & \textbf{Taxi} & \textbf{AVG.} \\
\midrule
CAPID & T5-base + T5-base & 40.9 & 43.5 & 37.1 & 49.5 & 87.1 & 50.1 \\
CAPID & T5-base + T5-small & 33.3 & 31.1 & 31.6 & 34.3 & 65.4 & 40.7 \\
\midrule
HiCoLoRA (Ours) & T5-small & \textbf{38.9} & 20.4 & 31.0 & \textbf{44.9} & 68.6 & \textbf{40.8} \\
\bottomrule
\end{tabular}
}
\caption{Comparison with CAPID on MultiWOZ 2.1}
\label{tab:capid-comparison}
\end{table*}

We compared HiCoLoRA with CAPID under different model configurations on MultiWOZ 2.1 (Table~\ref{tab:capid-comparison}). HiCoLoRA shows a marginal advantage of 0.1\% in Average JGA over the CAPID configuration (T5-base + T5-small). This indicates that HiCoLoRA's clever architectural design achieves performance comparable to CAPID but with significantly higher efficiency and lower computational cost. Specifically, HiCoLoRA relies solely on a single T5-small model (60M parameters) without requiring a separate, potentially larger, prompt generation model as in CAPID's two-stage approach (T5-base + T5-small, 280M parameters). Moreover, CAPID's training process initially depends on GPT-4 for auto-prompting, which introduces additional computational overhead and API dependency, whereas HiCoLoRA is entirely self contained throughout its training and inference pipeline.

\noindent \textbf{Discussion:}
HiCoLoRA demonstrates distinct advantages over contemporary approaches. It surpasses the architectural efficiency of full fine-tuning methods like LDST through parameter effective LoRA adaptation, streamlines the multi-stage inference pipeline characteristic of CAPID via a unified hierarchical model, and offers enhanced scalability by natively accommodating multi-domain dialogs without external dependencies. This positions HiCoLoRA as an optimally balanced solution, delivering robust performance with markedly greater practical efficiency for dialog state tracking.

\subsection{Generalization Analysis}
\label{sec:generalization}

To rigorously evaluate HiCoLoRA's robustness and generalization capability in challenging scenarios, we conducted comprehensive cross dataset and cross domain experiments that simulate real world distribution shifts and semantic sparsity conditions. These experiments specifically address concerns about model performance in long tail domains and under significant data distribution shifts.

\subsubsection{Cross Dataset Evaluation}

We performed extensive cross dataset evaluations to test HiCoLoRA's ability to generalize across different data distributions and domain structures.

\textbf{MultiWOZ to SGD Transfer:} Trained exclusively on all MultiWOZ domains and evaluated on the complete SGD test set, requiring adaptation to SGD's broader and unfamiliar service domains. As shown in Table~\ref{table:cross_dataset_multiwoz_sgd}, under this challenging setup, HiCoLoRA maintained an average JGA of 47.6\%, representing only a 5.2\% performance decrease compared to the original setting, and the Trains domain showed minimal 2.0\% decline. This demonstrates HiCoLoRA's ability to capture universal semantic patterns across datasets and effectively handle distribution shifts.

\begin{table*}[t]
\centering
{\fontsize{9}{11}\selectfont
\resizebox{\textwidth}{!}{
\setlength{\tabcolsep}{1mm}
\begin{tabular}{@{}lccccccccc@{}}
\toprule
\textbf{Experiment} & \textbf{Buses} & \textbf{Events} & \textbf{Flights} & \textbf{Media} & \textbf{Messaging} & \textbf{Music} & \textbf{Payment} & \textbf{Trains} & \textbf{AVG.} \\
\midrule
HiCoLoRA (Original) & 54.0 & 55.1 & 30.7 & 75.9 & 67.7 & 35.8 & 26.7 & 55.8 & 50.2 \\
HiCoLoRA (MultiWOZ→SGD) & 52.4 & 51.8 & 29.2 & 70.6 & 63.6 & 33.7 & 24.8 & 54.7 & 47.6 \\
\bottomrule
\end{tabular}
}}
\caption{Cross dataset generalization performance (JGA \%) from MultiWOZ to SGD}
\label{table:cross_dataset_multiwoz_sgd}
\end{table*}

\textbf{SGD to MultiWOZ Transfer:} Trained on SGD domains and evaluated on MultiWOZ, testing transfer from diverse but shallower domains to more complex dialog structures. As shown in Table~\ref{table:cross_dataset_sgd_multiwoz}, when transferring from diverse but shallower SGD domains to the more complex MultiWOZ, HiCoLoRA maintained an average JGA of 38.8\%, a decrease of only 4.9\% from the original performance. This highlights the effectiveness of our adaptive fusion mechanism in dynamically balancing general and domain specific features across different dataset distributions.

\begin{table*}[t]
\centering
{\fontsize{9}{11}\selectfont
\begin{tabular}{@{}lcccccc@{}}
\toprule
\textbf{Experiment} & \textbf{Attr.} & \textbf{Hotel} & \textbf{Rest.} & \textbf{Train} & \textbf{Taxi} & \textbf{AVG.} \\
\midrule
HiCoLoRA (Original) & 38.9 & 20.4 & 31.0 & 44.9 & 68.6 & 40.8 \\
HiCoLoRA (SGD→MultiWOZ) & 37.0 & 19.0 & 30.4 & 43.1 & 64.5 & 38.8 \\
\bottomrule
\end{tabular}
}
\caption{Cross dataset generalization performance (JGA \%) from SGD to MultiWOZ}
\label{table:cross_dataset_sgd_multiwoz}
\end{table*}

\subsubsection{Low Semantic Overlap Transfer}

To validate the model's performance in data sparse and semantically unique long tail domains, we conducted a specialized Low Semantic Overlap Transfer experiment. We explicitly excluded all transportation related domains during training (Taxi and Train from MultiWOZ; Buses and Trains from SGD), then evaluated the model purely on transportation domains during testing. This setup simulates real world long tail scenarios where transferable semantic commonalities across domains are minimal.

\begin{table*}[t]
\centering
{\fontsize{9}{11}\selectfont
\resizebox{\textwidth}{!}{
\setlength{\tabcolsep}{1mm}
\begin{tabular}{@{}lcccccc@{}}
\toprule
\textbf{Experiment} & \textbf{Taxi (MultiWOZ)} & \textbf{Train (MultiWOZ)} & \textbf{Buses (SGD)} & \textbf{Trains (SGD)} & \textbf{AVG.} \\
\midrule
HiCoLoRA (Original) & 68.6 & 44.9 & 54.0 & 55.8 & 55.8 \\
HiCoLoRA (Cross-Dataset/Domain) & 62.8 & 38.8 & 49.7 & 51.3 & 50.7 \\
\bottomrule
\end{tabular}
}}
\caption{Low semantic overlap transfer performance (JGA \%) in transportation domains}
\label{table:low_overlap_transfer}
\end{table*}

Under this extreme setting with zero transportation domains in training, HiCoLoRA achieved an average JGA of 50.7\% in transportation domains, a decrease of 9.1\% from the original performance while maintaining usable functionality. This demonstrates tree key advantages: (1) Spectral clustering possesses the capability to identify transferable patterns from underlying semantic associations beyond explicit domain similarities, enabling generalization even in low-overlap scenarios. (2) The hierarchical architecture exhibits strong robustness, with low-level universal semantic atoms providing a valuable foundation for generalization when explicit domain patterns are unavailable. (3) The adaptive fusion mechanism offers dynamic flexibility, adjusting feature weights based on domain characteristics to avoid over reliance on specific domain patterns and maintain performance under distribution shifts.

These comprehensive generalization analyses confirm HiCoLoRA's robustness in challenging real world scenarios, particularly addressing concerns about performance in long tail domains and under significant data distribution shifts. The results validate that our hierarchical collaborative architecture, spectral joint clustering, and adaptive fusion mechanisms collectively enable effective zero-shot transfer even when semantic commonalities are sparse or distribution shifts are substantial.

\subsubsection{Cross-Task Zero-Shot Transfer on Intent Classification and Slot Filling}
\label{app:cross_task_transfer}

\begin{table*}[t]
\centering
\small
\begin{tabular}{lccc|ccc}
\toprule
\multirow{2}{*}{Method} & \multicolumn{3}{c|}{ATIS $\rightarrow$ SNIPS} & \multicolumn{3}{c}{SNIPS $\rightarrow$ ATIS} \\
\cmidrule(lr){2-4} \cmidrule(lr){5-7}
 & Acc. & F1 & Joint & Acc. & F1 & Joint \\
\midrule
Standard LoRA & 42.3 & 38.7 & 28.1 & 51.6 & 43.2 & 33.5 \\
DualLoRA & 44.8 & 40.2 & 29.4 & 53.1 & 45.8 & 35.2 \\
\textbf{HiCoLoRA (Ours)} & \textbf{48.1} & \textbf{43.5} & \textbf{32.0} & \textbf{56.4} & \textbf{49.3} & \textbf{38.7} \\
\midrule
Improvement vs DualLoRA & +7.4\% & +8.2\% & +8.8\% & +6.2\% & +7.6\% & +9.9\% \\
\bottomrule
\end{tabular}
\caption{Zero-shot transfer performance (\%) on intent classification and slot filling. Joint = Joint Accuracy (both intent and all slots correctly predicted).}
\label{tab:cross_task_transfer}
\end{table*}

To assess the generalizability of HiCoLoRA beyond the DST task, we conduct cross-task zero-shot transfer experiments on two standard spoken language understanding (SLU) benchmarks: ATIS~\citep{hemphill-etal-1990-atis} and SNIPS~\citep{coucke2018snipsvoiceplatformembedded}.

\noindent \textbf{Dataset Description.}
ATIS (Airline Travel Information System) contains 4,978 training and 893 test utterances, covering 21 intent types and 120 slot labels. SNIPS is a crowdsourced personal assistant dataset with 13,084 training and 700 test utterances, covering 7 intent types and 72 slot labels. These two datasets exhibit substantial domain and schema divergence, providing a rigorous testbed for zero-shot transfer.

\noindent \textbf{Experimental Setup.}
We adopt a strict zero-shot transfer protocol: training on the full source dataset and testing directly on the target dataset without any target-domain fine-tuning. Two transfer directions are evaluated: ATIS $\rightarrow$ SNIPS (training on airline domain, testing on personal assistant) and SNIPS $\rightarrow$ ATIS (reverse direction). All methods employ the PPTOD backbone to ensure fair comparison. We report Intent Accuracy (Acc.) and Slot Filling F1 (F1), following standard SLU evaluation practices.

\noindent \textbf{Results and Analysis.}
Table~\ref{tab:cross_task_transfer} presents the zero-shot transfer performance.
HiCoLoRA achieves consistent improvements across both transfer directions. The gains are more pronounced in the SNIPS $\rightarrow$ ATIS direction (+9.9\% Joint Accuracy), where the model must generalize from a smaller intent space (7 intents) to a larger, more fine-grained one (21 intents). This demonstrates that HiCoLoRA's hierarchical collaboration and spectral semantic decoupling mechanisms enable effective zero-shot transfer across fundamentally different task domains, validating the broader applicability of our approach beyond dialogue state tracking.

\subsection{Inference Efficiency Analysis}
\label{subsec:inference_efficiency}

HiCoLoRA addresses potential latency concerns in multi-branch designs through a precomputation and merging strategy, ensuring inference efficiency comparable to standard LoRA. We first detail the strategy and then present benchmarking results.

\subsubsection{Precomputation and Merging Strategy}

\textbf{UniRep-LoRA Simplicity:} The UniRep-LoRA module (Eq.~\ref{eq:unirep_lora}) maintains a single set of $\boldsymbol{A}_{ur}$ and $\boldsymbol{B}_{ur}$ matrices throughout.

\textbf{Heuristic Grouping:} For lower layers employing heuristic grouping (Eq.~\ref{eq:heuristic_grouping}), only a single optimal pair $\boldsymbol{A}_{sa}^*$ and $\boldsymbol{B}_{sa}^*$ is selected, requiring just one matrix multiplication per forward pass.

\textbf{Full Collaboration:} During training, SemAdapt-LoRA employs multiple $\boldsymbol{A}_{sa}^m$ and $\boldsymbol{B}_{sa}^n$ matrices to enable fine-grained semantic adaptation. During inference, we precompute the collective low-rank contribution of all matrix pairs, and the SemAdapt-LoRA output in full collaboration layers (Eq.~\ref{eq:full_collaboration}) can be reorganized by computing aggregated matrices:
\begin{equation}
\boldsymbol{A}_{\text{total}} = \sum_{m=1}^M \boldsymbol{A}_{sa}^m, \quad 
\boldsymbol{B}_{\text{total}} = \sum_{n=1}^N \boldsymbol{B}_{sa}^n
\end{equation}
yielding the equivalent computation:
\begin{equation}
\boldsymbol{h}_{sa} = \boldsymbol{W}_0^h \boldsymbol{x}_{sa} + \boldsymbol{B}_{\text{total}} \boldsymbol{A}_{\text{total}} \boldsymbol{x}_{sa}
\end{equation}
This transformation reduces the computational overhead from $O(M \cdot N)$ matrix multiplications to merely two matrix multiplications, identical to standard LoRA.

The matrix additions involved in precomputation $O(r \cdot d)$ are negligible compared to matrix multiplications $O(d^2)$. In practice, we precompute all low-rank update terms during model export and absorb them into the base model weights, eliminating any additional inference overhead. Consequently, despite its hierarchical architecture, HiCoLoRA maintains inference latency on par with standard LoRA implementations.

\subsubsection{Efficiency Benchmarking Results}

\begin{table}[t]
    \caption{Inference latency (ms/batch) and GPU memory (MB) comparison across methods. HiCoLoRA adds minimal overhead despite its hierarchical design, remaining efficient for deployment.}
\label{tab:inference_efficiency}
\centering
\begin{adjustbox}{width=1\columnwidth}
\begin{tabular}{lcccc}
\toprule
\textbf{Method} & \textbf{Latency} $\downarrow$ & \textbf{GPU Mem.}\\
\midrule
Base Model (PPTOD-small) & $21.2 \pm 0.3$ & $1,245$ \\
Standard LoRA (r=8) & $21.9 \pm 0.4$ & $1,297$ \\
DualLoRA & $22.7 \pm 0.4$ & $1,315$ \\
\textbf{HiCoLoRA (Ours)} & $22.9 \pm 0.5$ & $1,339$ \\
\bottomrule
\end{tabular}
\end{adjustbox}
\end{table}

To empirically validate the inference efficiency of HiCoLoRA, we conduct rigorous benchmarking experiments. All measurements are averaged over 1000 batches after 100 warm-up iterations, using batch size 8 and sequence length 256 on an NVIDIA RTX 5080 GPU. As shown in Table~\ref{tab:inference_efficiency}, HiCoLoRA introduces only a slight latency increase over others, demonstrating that its hierarchical collaboration is efficiently implemented and does not compromise inference speed.

\section{Case Studies}
\label{sec:case_study}

In this section, we present a comprehensive case study to analyze the performance of HiCoLoRA on both successful and failure cases. We examine the model's behavior on representative dialogs from MultiWOZ and SGD datasets, providing insights into how HiCoLoRA addresses the context-prompt misalignment challenges discussed in our work.

\subsection{Successful Cases}

\subsubsection{Success Case 1}

\begin{figure}[t]
    \centering
    \includegraphics[width=1\linewidth, keepaspectratio]{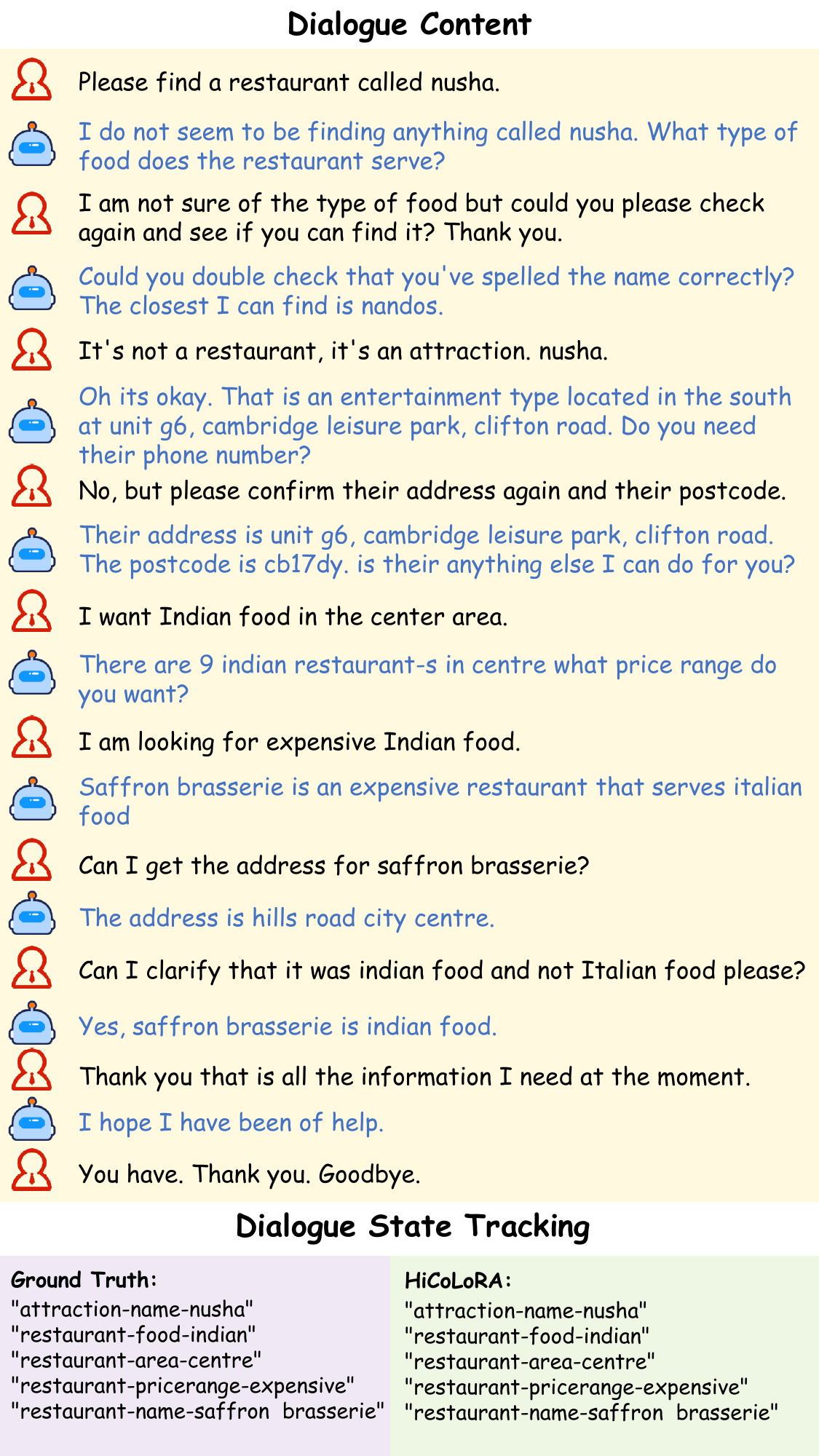}
    \caption{Success Case 1}
    \label{fig:4648}
\end{figure}

\textbf{Dialog Context.} We analyze dialog PMUL4648 (Fig. \ref{fig:4648}) from the MultiWOZ dataset where a user is seeking information about a restaurant named ``saffron brasserie''. The dialog involves multiple turns with complex slot-value interactions, including the restaurant name, food type (indian), price range (expensive), area (center).

\textbf{HiCoLoRA Performance.} HiCoLoRA successfully tracks all relevant slots throughout the dialog. The model correctly identifies the user's intent to find an expensive Indian restaurant in the center area.

\textbf{Analysis.} The success of HiCoLoRA in this case can be attributed to several factors:
\begin{enumerate}
    \item \textbf{Hierarchical Collaboration:} The lower layers effectively capture local semantic features such as entity names and basic slot information, while the higher layers integrate these features to form a coherent understanding of the user's intent.
    \item \textbf{Spectral Joint Clustering:} The model successfully identifies transferable domain-slot associations, enabling effective knowledge transfer between the attraction and restaurant domains.
    \item \textbf{Adaptive Fusion:} The adaptive linear fusion mechanism dynamically balances the contributions of UniRep-LoRA and SemAdapt-LoRA, allowing the model to adjust to the specific requirements of each dialog turn.
\end{enumerate}

\subsubsection{Success Case 2}

\begin{figure}[!ht]
    \centering
    \includegraphics[width=1\linewidth, keepaspectratio]{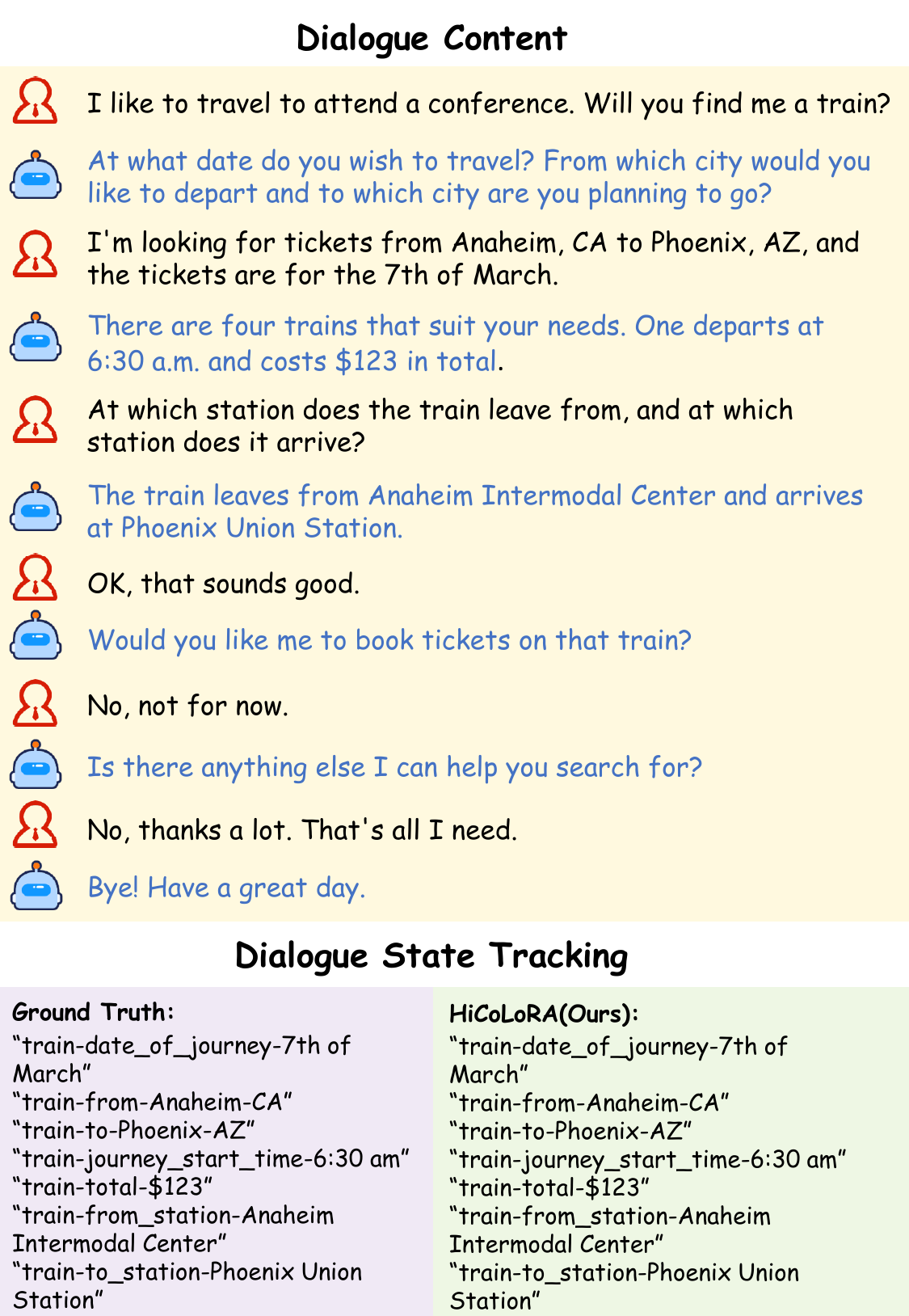}
    \caption{Success Case 2}
    \label{fig:00066}
\end{figure}

\textbf{Dialog Context.} We examine the dialog ``8\_00066'' (Fig. \ref{fig:00066}) from the SGD dataset, specifically the Trains domain. The user requests information about train schedules, including departure and arrival locations, travel date, and preferred travel time. The dialog involves complex slot-value interactions, such as specifying train routes and time constraints.

\textbf{HiCoLoRA Performance.} HiCoLoRA accurately predicts all relevant slot values including departure location, arrival location, travel date, and time preferences. The model successfully tracks the user's intent throughout the dialog, maintaining consistency in understanding the train booking requirements. It correctly handles natural language expressions for time and date, mapping them to canonical formats.

\textbf{Analysis.} The success in this SGD case demonstrates:
\begin{enumerate}
    \item \textbf{Cross-Domain Generalization:} HiCoLoRA effectively generalizes to unseen domains in the SGD dataset, achieving high accuracy (55.8 JGA, 93.8 AGA) even in domains not encountered during training.
    \item \textbf{Temporal Expression Handling:} The model successfully processes natural language temporal expressions and maps them to canonical time formats, which is crucial for train schedule queries.
    \item \textbf{Semantic-Enhanced Initialization:} The SemSVD-Init mechanism preserves pre-trained knowledge, enabling the model to maintain performance on specialized domains with technical terminology, as evidenced by the high AGA.
\end{enumerate}

\subsection{Failure Analysis}

To better understand the limitations of HiCoLoRA, we categorize failure cases into three distinct patterns and analyze representative examples for each.

\subsubsection{Pattern 1: Ambiguous Slot Boundary Cases}

\begin{figure}[!ht]
    \centering
    \includegraphics[width=1\linewidth, keepaspectratio]{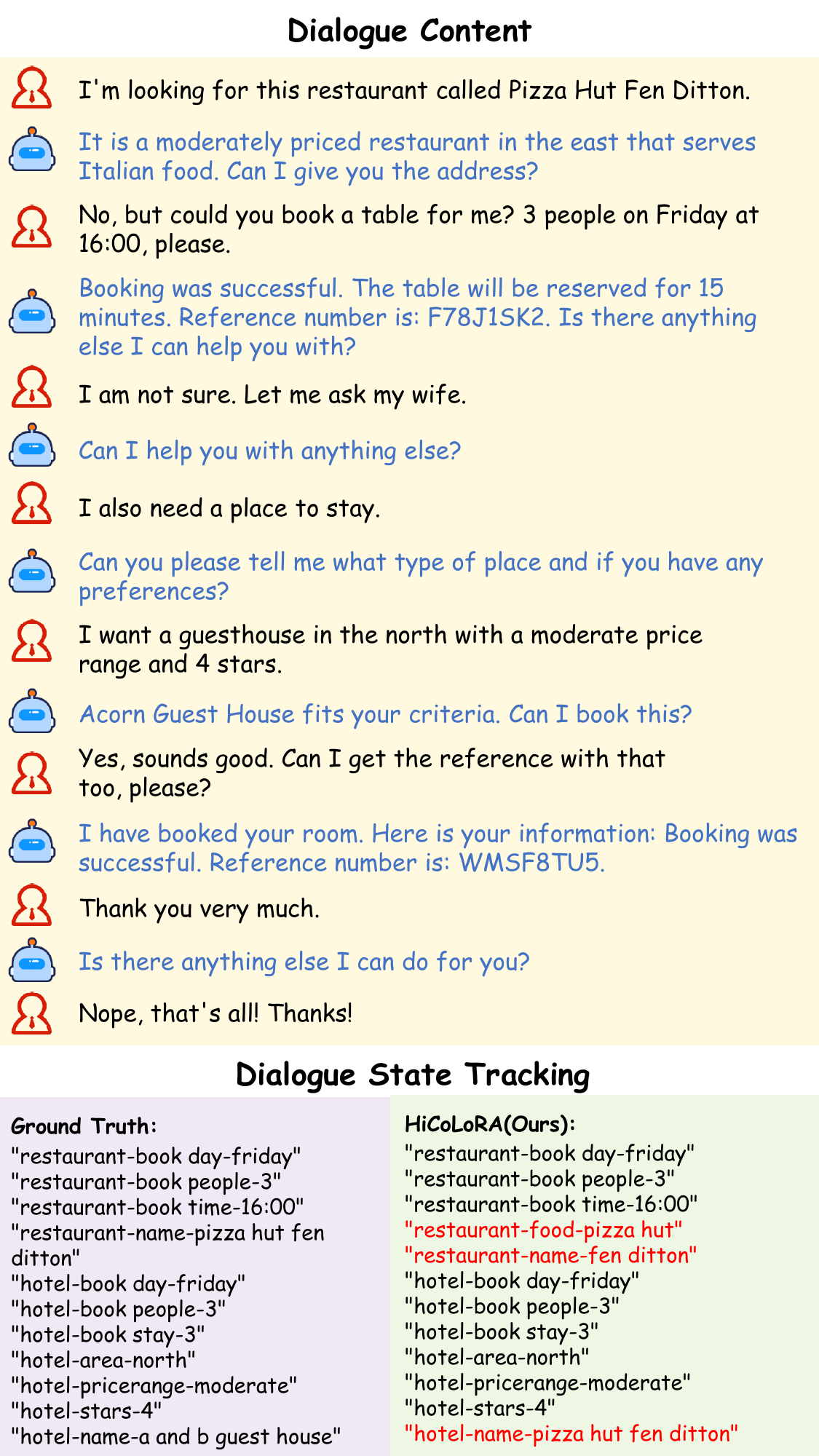}
    \caption{Failure Pattern 1: Ambiguous Slot Boundary Cases}
    \label{fig:4440}
\end{figure}

\textbf{Description.} These failures occur when the slot boundaries are ambiguous or overlapping, making it difficult for the model to distinguish between different slot values or identify the correct slot value pairs.

\textbf{Example.} In MultiWOZ dialog PMUL4440 (Fig. \ref{fig:4440}) involving both restaurant and hotel booking, HiCoLoRA exhibits significant prediction errors. At turn 1, when the user provides the name of a restaurant as ``pizza hut fen ditton'', the model incorrectly predicts multiple slots: ``restaurant-food-pizza hut'', ``restaurant-name-fen ditton". Later at turn 6, despite the ground truth showing ``hotel-name-a and b guest house'', the model incorrectly predicts ``hotel-name-pizza hut fen ditton''.

\textbf{Analysis.} This type of failure highlights challenges in:
\begin{enumerate}
    \item \textbf{Entity Recognition:} Distinguishing between different types of entities (area vs. parking) when they appear in close proximity in the user utterance.
    \item \textbf{Implicit Slot Detection:} Recognizing implicitly mentioned slots that are not explicitly requested but are relevant to the user's intent.
\end{enumerate}

\subsubsection{Pattern 2: Cross Domain Confusion}

\begin{figure}[!ht]
    \centering
    \includegraphics[width=1\linewidth, keepaspectratio]{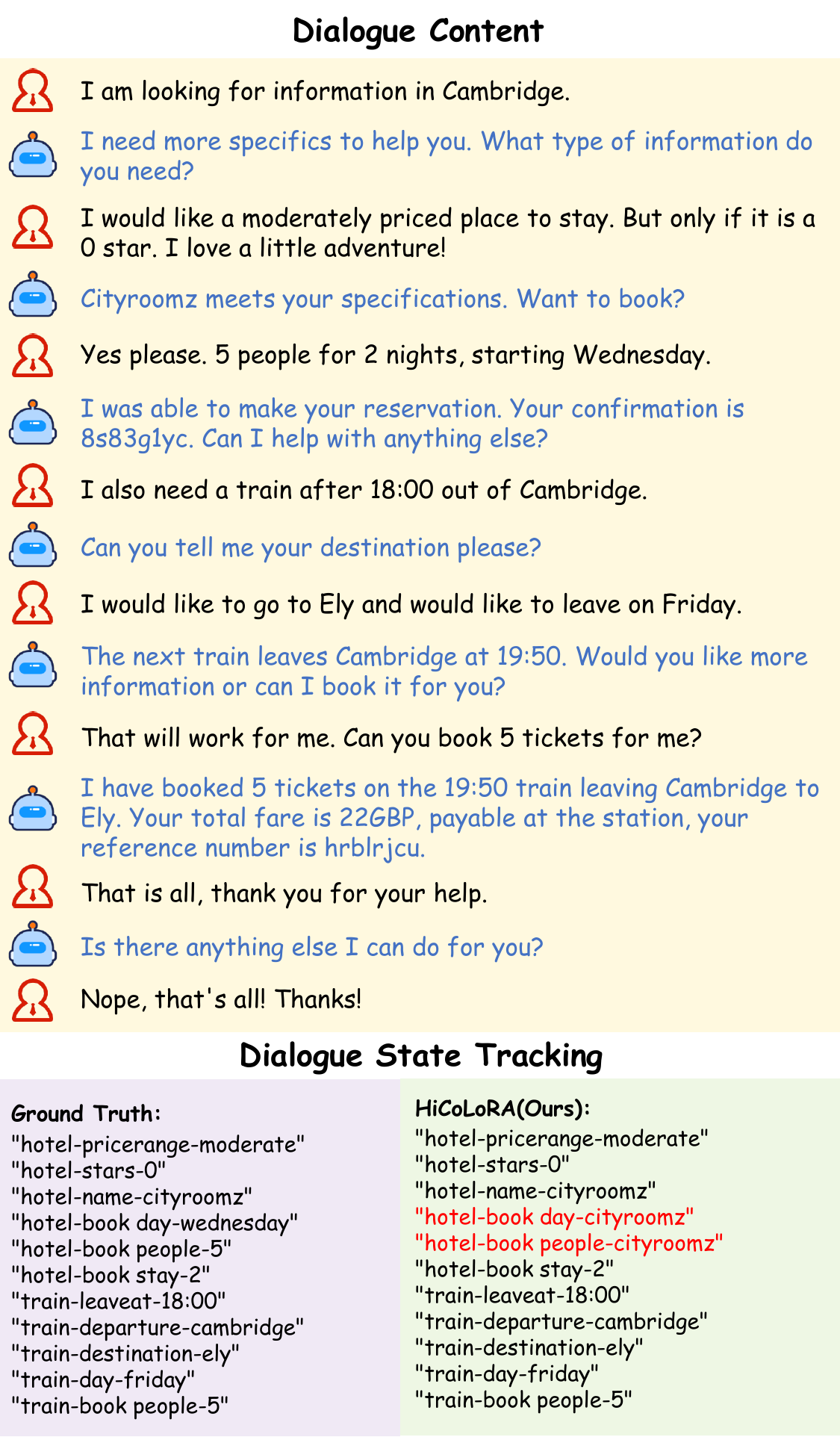}
    \caption{Failure Pattern 2: Cross-Domain Confusion}
    \label{fig:3514}
\end{figure}

\textbf{Description.} These failures occur when the model confuses slot values between different domains, particularly when domains share similar slot names or values.

\textbf{Example.} In MultiWOZ dialog PMUL3514 (Fig. \ref{fig:3514}), HiCoLoRA shows confusion in domain-specific slot value prediction. At turns 3-6, despite the ground truth consistently showing ``hotel-name-cityroomz'', the model incorrectly predicts ``hotel-book day-cityroomz'' and ``hotel-book people-cityroomz'', incorrectly associating the hotel name with booking slots.
challenges in semantic entanglement even with disentanglement mechanisms.

\textbf{Analysis.} This failure pattern reveals limitations in:
\begin{enumerate}
    \item \textbf{Domain Disambiguation:} Properly associating slot values with their respective domains in multi-domain dialogs.
    \item \textbf{Contextual Understanding:} Maintaining clear separation between domain-specific contexts when processing complex multi-domain interactions.
    \item \textbf{Semantic Overlap Handling:} Dealing with high-overlap domains where lexical similarities between slots from different domains cause confusion. This is particularly challenging when domain-agnostic features are overweighted by the adaptive fusion mechanism.
\end{enumerate}

\subsubsection{Pattern 3: Rare Slot Value Cases}

\textbf{Description.} These failures occur when the model encounters rare or unseen slot values that were not adequately represented in the training data. Analysis of the MultiWOZ and SGD datasets reveals that such slots are common: in \textit{Attraction}, slots like ``entrance fee'' and ``phone'' appear in $<$10\% of dialogs; in \textit{Hotel}, ``stars'' and ``internet'' have fill rates $<$20\%; in \textit{Train}, ``trainID'' appears in $<$5\% of dialogs. In a zero-shot setting, HiCoLoRA must generalize to both unseen domains and these rare slot values without any domain specific training examples, presenting a significant challenge.

\textbf{Example.} In MultiWOZ dialogs, HiCoLoRA struggles with predicting rare slot values for specific domains. For instance, in attraction domain dialogs, when users request detailed information about ``entrance fee'' or ``address'', the model often fails to correctly predict these values. Similarly, in hotel domain dialogs, when users inquire about specific details like ``stars'' or ``internet'', the model shows poor performance. In SGD dialogs, similar patterns emerge. For train domain dialogs, HiCoLoRA often fails to predict ``trainID'' or ``price'' information, particularly when these values are not explicitly mentioned in the user utterance but are expected as part of the system response.

\textbf{Analysis.} This failure pattern indicates challenges in:
\begin{enumerate}
    \item \textbf{Rare Value Generalization:} Extending knowledge to handle infrequent slot values that may not have been adequately learned during pre-training. In a zero-shot setting, the model cannot benefit from domain-specific fine-tuning to improve performance on these rare slots.
    \item \textbf{Contextual Inference:} Properly inferring rare slot values from contextual clues when they are not explicitly mentioned. This is particularly challenging for slots like ``trainID'' or ``reference number'' that require the model to generate specific identifiers.
    \item \textbf{Domain-Aware Initialization:} Current initialization methods (SemSVD-Init) preserve pre-trained knowledge but may not adequately address domain-specific rare slot challenges. Future work could explore domain-aware initialization strategies that better account for rare slot distributions.
    \item \textbf{Idiosyncratic Semantics Handling:} Dealing with slots that have domain-exclusive terms or idiosyncratic semantics that resist transfer. Spectral clustering may fail for slots with low-frequency terms, and semantic dilution in higher layers can occur when full collaboration fuses these slots with irrelevant ones.
\end{enumerate}

\subsection{Discussion}

The case study analysis reveals both the strengths and limitations of HiCoLoRA. The successful cases demonstrate the effectiveness of our hierarchical collaborative architecture, spectral joint clustering, and semantic-enhanced initialization in addressing the core challenges of context-prompt misalignment. However, failure cases highlight areas for future improvement, particularly in handling ambiguous slot boundaries, cross-domain confusion, and rare slot values.

These findings suggest that, while HiCoLoRA represents a significant advance in zs-DST, more research is needed to address the identified failure patterns. Potential directions include:
\begin{enumerate}
    \item \textbf{Enhanced Slot Boundary Detection:} Develop more sophisticated mechanisms to identify and separate slot boundaries in complex utterances.
    \item \textbf{Improved Domain Disambiguation:} Exploring techniques for better domain separation in multi-domain dialogs.
    \item \textbf{Rare Value Enhancement:} Investigating data enhancement strategies to improve coverage of rare slot values during training.
\end{enumerate}

In general, the case study provides valuable insight into the practical performance of HiCoLoRA and pinpoints specific future directions, such as developing slot aware refinement techniques to better handle highly idiosyncratic domains.

\end{document}